\documentclass{article}

\PassOptionsToPackage{numbers, compress}{natbib}

\usepackage[preprint]{neurips_2025}

\usepackage[utf8]{inputenc} 
\usepackage[T1]{fontenc}    
\usepackage{hyperref}       
\usepackage{url}            
\usepackage{booktabs}       
\usepackage{amsfonts}       
\usepackage{nicefrac}       
\usepackage{microtype}      
\usepackage{xcolor}         
\usepackage{amsmath}
\usepackage{array, ragged2e}
\DeclareMathOperator*{\argmax}{arg\,max}
\usepackage{graphicx}
\newcolumntype{P}[1]{>{\raggedright\arraybackslash}p{#1}}
\usepackage{amssymb}
\usepackage{enumitem}
\usepackage{pifont}
\usepackage{wrapfig}
\usepackage{tikz}
\usepackage{multirow}
\usetikzlibrary{arrows.meta, positioning, bending}  
\usepackage{tikz-cd}
\usepackage{subcaption}

\usepackage{algorithm}
\usepackage{algpseudocode}

\newtheorem{remark}{Remark}

\title{ 
QueryBandits for Hallucination Mitigation: 

Exploiting Semantic Features for No-Regret Rewriting}

\author{%
  Nicole Cho, William Watson, Alec Koppel, Sumitra Ganesh, Manuela Veloso \\
  JP Morgan AI Research \\
  New York, NY \\
  \texttt{nicole.cho@jpmorgan.com} \\
}

\begin{document}

\maketitle

\begin{abstract}
\label{sec: abstract}
Advanced reasoning capabilities in Large Language Models (LLMs) have caused higher hallucination prevalence; yet most mitigation work focuses on after-the-fact filtering rather than shaping the queries that trigger them. 
We introduce \textbf{QueryBandits}, a bandit framework that designs rewrite strategies to maximize a reward model, that encapsulates hallucination propensity based upon the sensitivities of 17 linguistic features of the input query---and therefore, proactively steer LLMs away from generating hallucinations.
Across 13 diverse QA benchmarks and 1,050 lexically perturbed queries per dataset, our top contextual QueryBandit (Thompson Sampling) achieves an 87.5\% win rate over a no-rewrite baseline and also outperforms zero-shot static prompting ("paraphrase" or "expand") by 42.6\% and 60.3\% respectively. 
Therefore, we empirically substantiate the effectiveness of QueryBandits in mitigating hallucination via the intervention that takes the form of a query rewrite. Interestingly, certain static prompting strategies, which constitute a considerable number of current query rewriting literature, have a higher cumulative regret than the no-rewrite baseline, signifying that static rewrites can worsen hallucination. Moreover,  we discover that the converged per-arm regression feature weight vectors substantiate that there is \textit{no} single rewrite strategy optimal for all queries. In this context, guided rewriting via exploiting semantic features with QueryBandits can induce significant shifts in output behavior through forward-pass mechanisms, bypassing the need for retraining or gradient-based adaptation.

\begin{figure}[t]
  \centering
  \includegraphics[width=\textwidth, trim=0.34cm 6.6cm 0.4cm 3.5cm, clip]{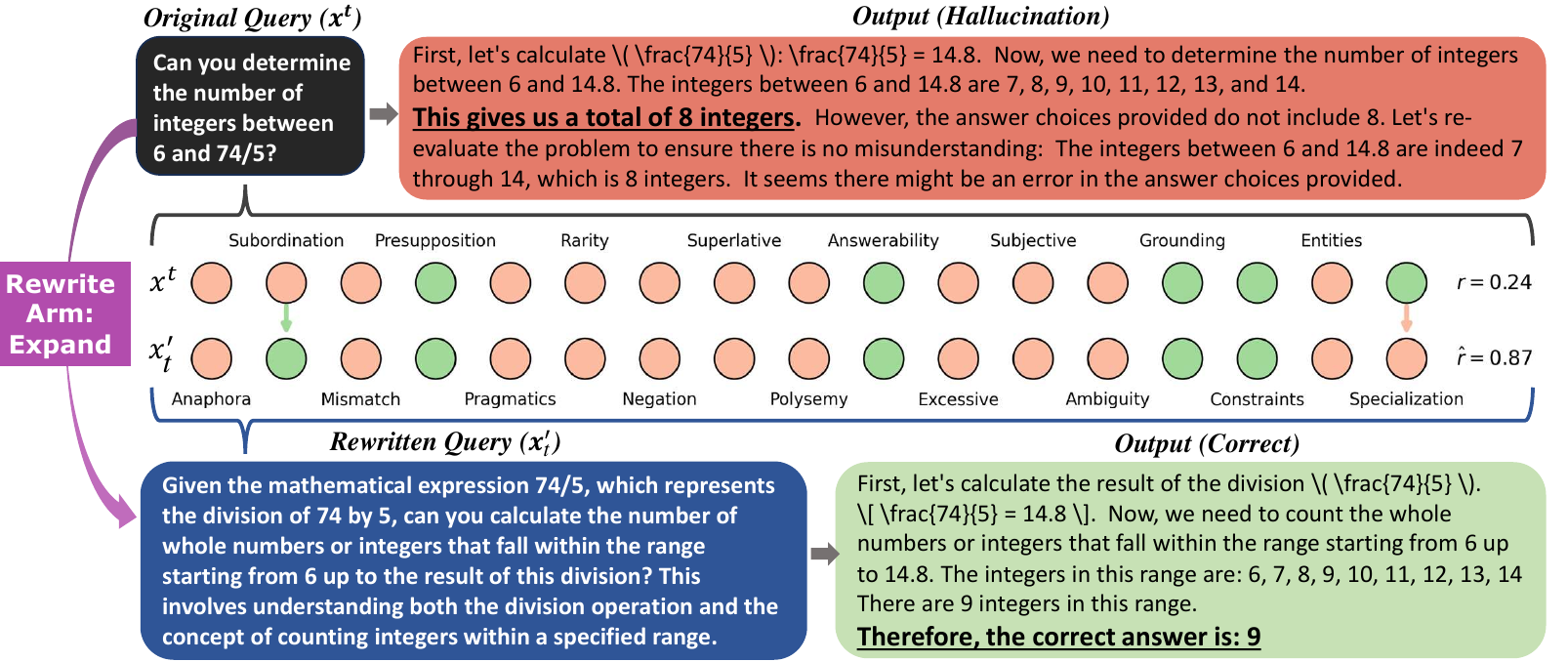}
    \caption{\textbf{QueryBandits and Its Success in Mitigating Hallucination.} The original query \(x_t\) induces a hallucinatory output: the LLM calculates 8 integers between 6 and 74/5. QueryBandits, by leveraging the feature vector, selects the \textsc{Expand} rewrite strategy. The rewritten query \(x'_t\) generates an accurate output of 9 integers. Noticeably, the feature vectors are different in the rewrite \(x'_t\) - \textit{subordination (more complex clauses)} is now present while \textit{specialization (query requiring domain-specific knowledge for understanding)} is absent - signifying effects of the \textsc{Expand} strategy.}
    \label{fig:anecdote}
    \vspace{-5mm}
\end{figure}

\end{abstract}

\section{Introduction}
\label{sec: Introduction}
As Large Language Models (LLMs) grow more powerful, their hallucinations increase in severity \citep{OpenAI2025, NYTimes2025AIHallucinations}. Hallucinations refer to the generation of inaccurate outputs, according to the LLM's internal "understanding" of the query \citep{ji2023mitigatinghallucinationlargelanguage}. Mitigation efforts, however, remain largely confined to retrofitting the outputs instead of reshaping the query \citep{ ji2023mitigatinghallucinationlargelanguage, tonmoy2024comprehensivesurveyhallucinationmitigation}, even though LLM outputs are highly variable to lexical perturbations of the incoming query \citep{Watson_Cho_Srishankar_2025, cho2025multiqaanalysismeasuringrobustness}. This style of analysis is in line with recent trends of mechanistic interpretability \citep{ameisen2025circuit}, where interpretable subgraphs of neural architectures are sought on the internal embeddings. 
However, these subgraphs may be unavailable due to enterprise restrictions \citep{touvron2023llama2openfoundation}; moreover, there is no clear way to link a particular subgraph template with hallucination. 

In this work, we design interventions to mitigate hallucinations based on semantic features of a query. To formalize the relationship between a query and hallucination, we associate with each query   the output's propensity to be hallucinatory, which is implicitly associated with a reward model. We further hypothesize that this distribution is a function of a potential intervention which takes the form of a query rewrite. To be more precise, 
for \( s_{\text{llm}} \in [0,1] \), a binary score assigned by an LLM-based judge \citep{liu2023gevalnlgevaluationusing,
adlakha2024evaluatingcorrectnessfaithfulnessinstructionfollowing}, 
\( s_{\text{fuzz}} \in [0,1] \), a fuzzy string similarity metric \citep{max_bachmann_2024_10938887}, and \( s_{\text{bleu}} \in [0,1] \), the BLEU-1 score capturing unigram lexical overlap \citep{Papineni02bleu:a, lin-och-2004-orange, callison-burch-etal-2006-evaluating},
we define hallucination in terms of a reward model \(
r_t = \alpha \cdot s_{\text{llm}} + \beta \cdot s_{\text{fuzz}} + \gamma \cdot s_{\text{bleu}}
\) where hallucinatory responses are those associated with low rewards. Through our ablations, we also discover the Pareto-optimal balance of weights (Fig.~\ref{fig:simplex}), by assigning a higher weight on LLM-as-a-judge, which is also evidenced in studies that highlight the efficacy of LLMs in Natural Language Generation (NLG) evaluation tasks \citep{wang-etal-2023-chatgpt, yuan2021bartscoreevaluatinggeneratedtext, fu2023gptscoreevaluatedesire}. 
As we are agnostic to the stationarity of the reward distribution or lack thereof due to the extreme dimensionality of the output space \citep{riemer2022continual, ji-etal-2023-towards}, we propose to evaluate whether rewrite strategies exhibit advantages when seeking to optimize the average reward or its worst case. 

 The usage of reinforcement learning (RL) \cite{sutton2018} methods have been applied in Natural Language Processing (NLP) tasks such as optimizing document-level query search \citep{nogueira-cho-2017-task}, fine-tuning LLMs \citep{deepseekai2025deepseekr1incentivizingreasoningcapability,ouyang2022traininglanguagemodelsfollow}, and post-training \citep{mudgal2024controlleddecodinglanguagemodels}. Despite its prevalent usage, to our knowledge, there is no in-depth interactive rewriting research to mitigate hallucination. We focus on bandit based methods because: (i) modeling the long-term value of hallucination manifestation would require multiple queries from a common sub-population; (ii) averaging hallucination propensity across distinct contexts may obscure per-query contextual idiosyncrasies; and (iii) the token concatenation that defines how vocabulary sampling occurs in output generation is \emph{deterministic}, meaning it is unclear if an MDP transition model may even be defined. That is not to say bandit methods have no precedent in NLP. Proximal Policy Optimization \citep{schulman2017proximalpolicyoptimizationalgorithms} variants for LLMs such as GRPO (Group Relative Policy Optimization) \citep{
shao2024deepseekmathpushinglimitsmathematical} and ReMax \citep{li2024remaxsimpleeffectiveefficient} also remove the critic via grouped Monte Carlo or baseline-adjusted returns. 

We have meticulously selected five distinct rewriting arms/strategies; to our knowledge, current research tends to pursue a "one-size-for-all" approach, leveraging one of these rewriting strategies for \textit{all} query types, and does not pursue guided rewrites via bandits \citep{ma2023queryrewritingretrievalaugmentedlarge, Watson_Cho_Srishankar_2025}. We also have rigorously selected 17 different linguistic features that are known to hinder/enhance human or LLM understanding (Table~\ref{tab:feature_vector_citations}). We frame query rewriting as an online decision problem and by leveraging per-query features, our bandit framework, \textbf{QueryBandits}, allocates exploration where uncertainty is high and exploitation where features are meaningful---resulting in hallucination reduction.

{\noindent \bf Contribution 1:}
We introduce an empirically validated reward function, combining an LLM‐judge, fuzzy‐match, and BLEU metrics (with \(\alpha,\beta,\gamma=(0.6,0.3,0.1)\)) chosen inside the 1\% Pareto‐optimal frontier on a held‐out human‐labeled set, to drive context‐aware bandit learning (Fig.~\ref{fig:simplex}). Guided by this reward signal, our contextual QueryBandits learn to tailor each query rewrite to its linguistic/contextual fingerprint. Our best performing contextual bandit, QueryBandits-Thompson Sampling, drives a 87.5\% win-rate boost over the \textsc{No-Rewrite} baseline---a considerable margin that highlights the efficacy of rewriting in reducing hallucination. 

{\noindent \bf Contribution 2:}
QueryBandits-Thompson Sampling delivers a decisive 42\% gain against the predominant static prompting strategy (\textsc{Paraphrase}), underscoring that feature-aware rewriting with bandits is effective for mitigating hallucination. In Figure~\ref{fig:excess_regret}, it is clear that the contextual QueryBandits hone into the optimal rewrite quickly, accruing an order-of-magnitude less cumulative regret than static prompting, vanilla (non-contextual) bandits, or no-rewriting. These gains confirm that a feature-aware, online adaptation mechanism can consistently outpace one-shot heuristics in mitigating hallucinations. An interesting finding is that some static prompting methods 
have a higher cumulative regret than \textsc{No-Rewrite}, demonstrating that zero-shot prompting can cause more severe hallucinations. 

{\noindent \bf Contribution 3:} 
We provide empirical evidence that there is no \textit{single} rewrite strategy that maximizes the reward for all types of queries. Our analysis of the per-arm regression weights (Figure~\ref{fig:raw-feature-contrib}) reveals how each arm’s effectiveness hinges on the semantic features of a query. For example, if a query displays the feature \emph{(Domain) Specialization}, meaning that the query can only be understood with domain-specific knowledge, the rewrite arm \textsc{Expand} is very effective in contrast to \textsc{Simplify} (Figure~\ref{fig:anecdote}). Crucially, ablating the 17-dimensional feature input causes QueryBandits-Thompson Sampling’s performance to drop to just 81.7\% win rate and 754.66 exploration-adjusted reward. This performance gap confirms that the linguistic features carry an associative signal about optimal rewrite strategy. Finally, we observe that across datasets, higher feature variance coincides with greater variance in arm selection (Figure~\ref{fig:rel-feature-contrib}), resulting in genuinely diverse arm choices (Figure~\ref{fig:rank}). 

{\noindent \bf Contribution 4:} 
Optimizing queries post-training by embedding them directly into the stage of prompting with minimal computational or token overhead constitutes an efficient strategy for trustworthy interfacing with LLMs, particularly under resource-constrained or latency-sensitive conditions. We bypass the need for retraining or gradient-based adaptation through purely forward-pass mechanisms. Moreover, through QueryBandits, we provide a mechanism to interpret the sensitivity of LLM performance to contextual rewrites. 

{\noindent \bf Interesting Findings:} 
On many standard benchmark datasets, we discover that linear contextual bandits converge almost exclusively to the \textsc{No Rewrite} arm (Figure~\ref{fig:no-rewrite}), empirically exposing LLM's tendency for query memorization on benchmarks. Only when we introduce semantically invariant but lexically perturbed queries does the policy meaningfully diversify across rewrite strategies; a meaningful insight for the research community that surface-form novelty is essential in training query rewriting algorithms. On the non-contextual bandits, we empirically discover that they converge to a single rewrite strategy within a dataset, in contrast to contextual bandits that tend to diversify its choices conditioned on the presence/absence of linguistic features.


\section{Related Work}\label{sec:related}

\begin{table}[t]
\small
\centering
\caption{Binary linguistic feature vector $\mathbf{f} \in \{0,1\}^{17}$ identified as challenging from a linguistics and LLM perspective. Features are grouped by type and grounded in prior work. For more specific examples, see Appendix Table \ref{tab:feature_examples}.} 
\label{tab:feature_vector_citations}
\resizebox{\textwidth}{!}{%
\begin{tabular}{llc}
\toprule
\textbf{Feature} & \textbf{Description} & \textbf{Citation} \\
\midrule
\multicolumn{3}{l}{\textit{Structural Features}} \\
\midrule
Anaphora & Contains anaphoric references (e.g., \textit{it}, \textit{this}) & \citep{schuster-1988-anaphoric, chen-etal-2018-preco} \\
Subordination & Contains multiple subordinate clauses & \citep{jeong-etal-2024-adaptive, alajrami2022doespretrainingobjectiveaffect, blevins2023promptinglanguagemodelslinguistic} \\
\midrule
\multicolumn{3}{l}{\textit{Scenario-Based Features}} \\
\midrule
Mismatch & Query (e.g. open-ended) does not match task (e.g. retrieval) & \citep{ gao2024retrievalaugmentedgenerationlargelanguage, Kamath_2024} \\
Presupposition & Assumptions within the query are implicitly regarded as truthful & \citep{Karttunen2016, Levinson1983, VanDerSandt1992} \\
Pragmatics & Queries with discourse-driven intent (i.e. "can you pass me the salt") & \citep{sravanthi-etal-2024-pub, Levinson1983, Sadock-and-Zwicky-1985} \\
\midrule
\multicolumn{3}{l}{\textit{Lexical Features}} \\
\midrule
Rarity & Presence of rare words with poor representation & \citep{schick2019rarewordsmajorproblem, Khassanov2019} \\
Negation & Presence of negation  (e.g., \textit{not}, \textit{never}) & \citep{hossain2022leveragingaffirmativeinterpretationsnegation, kassner-schutze-2020-negated, truong2023languagemodelsnaysayersanalysis} \\
Superlative & Usage of superlative forms (e.g., \textit{best}, \textit{largest}) with implicit semantics & \citep{pyatkin2024superlativescontextmodelingimplicit, Farkas} \\
Polysemy & Presence of words that have multiple, related-meanings & \citep{ansell2021polylmlearningpolysemylanguage, haber-poesio-2024-polysemy} \\
\midrule
\multicolumn{3}{l}{\textit{Stylistic Complexity}} \\
\midrule
Answerability & Query is not highly speculative, sarcastic or rhetorical & \citep{10.1609/aaai.v37i8.26138,belfathi2023harnessinggpt35turborhetoricalrole, lewis2021retrievalaugmentedgenerationknowledgeintensivenlp} \\
Excessive & Overloaded with a large amount of details and information & \citep{li2024longcontextllmsstrugglelong, liu2023lostmiddlelanguagemodels} \\
Subjectivity & Query requires LLM to reflect creatively and engender a personal opinion & \citep{durmus2024measuringrepresentationsubjectiveglobal, lv-etal-2024-subjective} \\
Ambiguity & Presence of ambiguous phrasing that opens multiple interpretations & \citep{brown2020languagemodelsfewshotlearners, kim-etal-2024-aligning, liu-etal-2023-afraid} \\
\midrule
\multicolumn{3}{l}{\textit{Semantic Grounding}} \\
\midrule
Grounding & Presence of clear intention and goal & \citep{Clarke, wei2023chainofthoughtpromptingelicitsreasoning} \\
Constraints & Presence of temporal/spatial/task-specific constraints & \citep{jiang2024followbenchmultilevelfinegrainedconstraints,  lewis2021retrievalaugmentedgenerationknowledgeintensivenlp} \\
Entities & Presence of verifiable entities & \citep{lee2023factualityenhancedlanguagemodels, ashok2023promptnerpromptingnamedentity, wang2023gptnernamedentityrecognition} \\
Specialization & Query requires domain-specific knowledge for understanding & \citep{watson-etal-2025-law, Cho_2024, zeng2024flowmindautomaticworkflowgeneration} \\
\bottomrule
\end{tabular}
}
\vspace{-0.5cm}
\end{table}

 
LLM Hallucinations are known to erode trustworthiness from a societal perspective \citep{dechert2024ai}. Recently, certain conceptual analyses frame it as a new epistemic failure mode, requiring dedicated mitigation agendas \citep{yao2024llmlieshallucinationsbugs, ouyang2022traininglanguagemodelsfollow}. Moreover, the release of more advanced reasoning models are concerningly generating more hallucinations, as reported in OpenAI's technical report \citep{OpenAI2025} on o3 and o4-mini. The \citet{NYTimes2025AIHallucinations} recently reported real-world case studies on how fabricated LLM outputs are prompting legal accountability. Especially, the advent of more LLM-Agent enabled systems \citep{watson-etal-2025-law, watson-etal-2023-hiddentables} will engender the compounding cost of hallucinations. 

Thus, mitigation is  regarded as indispensable for faithful LLM interactions \citep{ji2023mitigatinghallucinationlargelanguage, tonmoy2024comprehensivesurveyhallucinationmitigation, Huang_2025} and research is expanding from post-hoc detection \citep{ madaan2023selfrefineiterativerefinementselffeedback}, to preemptive grounding. 
Parallel to human‐in‐the‐loop RLHF, RL from AI Feedback (RLAIF) trains a reward model on preferences generated by an LLM—bypassing expensive human labels—while exceeding RLHF quality on summarization and dialogue tasks \citep{lee2024rlaifvsrlhfscaling, ouyang2022traininglanguagemodelsfollow,
christiano2023deepreinforcementlearninghuman}. \citet{Watson_Cho_Srishankar_2025} introduced pre-generation hallucination estimation via query perturbations. \citet{ma2023queryrewritingretrievalaugmentedlarge} has proposed the \textit{Rewrite-Retrieve-Read} framework for Retrieval Augmented Generation pipelines while human-designed query rules has been heavily used for rewriting \citep{liu2024queryrewritinglargelanguage, mao-etal-2024-rafe, chen-etal-2024-prompt}.
A common theme is that either raw prompting or  manual heuristics is used - not guided rewrites, through contextual signals of the original query.

\citet{blevins2023promptinglanguagemodelslinguistic} has conducted extensive research on the strong performance of Pretrained Language Models (PLMs) to retrieve linguistic features of a query in a few-shot manner. We employ this research and leverage an LLM to identify key linguistic features as outlined in Table \ref{tab:feature_vector_citations}. For the selection of these features, we have thoroughly reviewed not only existing LLM literature but also traditional linguistics to identify features that are known to impact both human and LLM understanding. 

%


\begin{table}[!t]
    \centering
    \caption{Overview of datasets, including domain, license, number of examples, associated scenarios, etc. These datasets span a diverse range of question types, domains, and reasoning skills, supporting robust evaluation. E = Extractive, M = Multiple Choice, A = Abstractive.}
    \label{tab:dataset_details}
    \small
    \resizebox{0.9\textwidth}{!}{%
    \begin{tabular}{lcccccc}
        \toprule
        \textbf{Dataset} & \textbf{Scenario} & \textbf{Domain} & \textbf{License} & \textbf{Count} & \textbf{Citation} \\
        \midrule
        SQuADv2         & E, A   & Wikipedia              & CC BY-SA 4.0 & 86K  & \citep{rajpurkar2016squad, rajpurkar2018know} \\
        TruthfulQA      & M, A   & General Knowledge      & Apache-2.0   & 807  & \citep{lin-etal-2022-truthfulqa} \\
        SciQ            & M, A   & Science                & CC BY-NC 3.0 & 13K  & \citep{SciQ} \\
        MMLU            & M      & Various                & MIT          & 15K  & \citep{hendryckstest2021} \\
        PIQA            & M      & Physical Commonsense   & AFL-3.0      & 17K  & \citep{Bisk2020} \\
        BoolQ           & M      & Yes/No Questions       & CC BY-SA 3.0 & 13K  & \citep{clark2019boolq, wang2019superglue} \\
        OpenBookQA      & M      & Science Reasoning      & Apache-2.0   & 6K   & \citep{OpenBookQA2018} \\
        MathQA          & M      & Mathematics            & Apache-2.0   & 8K   & \citep{amini-etal-2019-mathqa} \\
        ARC-Easy        & M      & Science                & CC BY-SA 4.0 & 5K   & \citep{allenai:arc} \\
        ARC-Challenge   & M      & Science                & CC BY-SA 4.0 & 2.6K & \citep{allenai:arc} \\
        WikiQA          & A      & Wikipedia QA           & Other        & 1.5K & \citep{yang-etal-2015-wikiqa} \\
        HotpotQA        & A      & Multi-hop Reasoning    & CC BY-SA 4.0 & 72K  & \citep{yang-etal-2018-hotpotqa} \\
        TriviaQA        & A      & Trivia                 & Apache-2.0   & 88K  & \citep{2017arXivtriviaqa} \\
        \bottomrule
    \end{tabular}%
    }
\end{table}

\section{Methodology and Evaluation Metrics}\label{sec:methodology}

In this section, we define the key ingredients to formulate a sequential decision-making problem as a multi-armed bandit \cite{lattimore2020bandit}. Specifically, we define the action space, contextual attributes, and reward. 

{\noindent \bf Bandit Formulation.} 
In the contextual bandit framework \cite{lattimore2020bandit}, a learner is faced with, given a context vector $x_t\in\mathcal{X}\subset\mathbb{R}^d$ at time $t$, selecting an arm $a_t$ from an action set $\mathcal{A}$. Upon that basis, Nature reveals a reward $r_t(x_t,a_t)$ which is a function of the context and arm. To be more precise, the reward is defined as $r: \mathcal{X}\times \mathcal{A} \rightarrow \mathbb{R}$. The goal of a bandit algorithm is to select arms that are eventually good with respect to the average (or cumulative) reward. In the stochastic bandit setting, specifically, one is interested in selecting arms according to a \emph{policy} $\pi: \mathcal{X} \rightarrow \rho(\mathcal{A})$ which performs as well as $\max_{\pi\in\Pi} \mathbb{E}[r(x,\pi(x))]$. Here $\rho(\mathcal{A})$ denotes the probability simplex over $K$ arms, and $\Pi$ is some class of policies. Next we specify the concrete choices of action space, context variables, and rewards for the query rewrite setting.

{\noindent \bf Action Space.} Let $\mathcal{A} = \{a_0, a_1, a_2, a_3, a_4\}$ denote the set of rewriting strategies (arms), where each arm represents a distinct style of query reformulation implemented via prompt-based instructions to an LLM. As outlined in \S\ref{sec:related}, previous research tends to take a "one-size-for-all" approach, 
\begin{itemize}[noitemsep, leftmargin=*, topsep=0pt, partopsep=0pt, label={\tiny\raisebox{0.5ex}{$\blacktriangleright$}}]

    \item $a_0$: \textbf{Paraphrasing} - The incoming query is rewritten using a prompt such as “Paraphrase this question while preserving its meaning.” This introduces lexical diversity while maintaining semantic similarity, testing whether alternative phrasings reduce hallucination. Many have explored how paraphrasing can improve factual consistency in LLMs \citep{cho2025multiqaanalysismeasuringrobustness,deng2024rephraserespondletlarge, Witteveen_2019} .  
    \item $a_1$: \textbf{Simplification} - The incoming query is rewritten to eliminate nested clauses or complex syntax.  This targets hallucinations caused by long-range dependencies or overloaded details - and borrows ideas from educational psychology where simpler, granular, prompts enable a child to learn a new skill \citep{libby2008}. Recently, \citet{van2021ihelpyouusing, zhou2023leasttomostpromptingenablescomplex} report on how simplified prompts decrease off-topic generations and enable complex reasoning tasks.
    \item $a_2$: \textbf{Disambiguation} - The query is rewritten by disambiguating vague references (e.g., ambiguous pronouns or temporal expressions). Many have conducted extensive studies on LLMs' inabilities to resolve ambiguous queries which leads to subpar performance \citep{
    deng-etal-2023-prompting, shahbazi2019entityawareelmolearningcontextual, cole-etal-2023-selectively}. 
    \item $a_3$: \textbf{Expansion} - The query is rewritten to explicitly expand on relevant named entities or attributes, elaborating on contextual cues through generation \citep{yu2023generateretrievelargelanguage}. Since transformers-based LLMs optimize next-token likelihood over attention-mediated context windows \citep{vaswani2023attentionneed}, appending fine-grained query constraints effectively conditions the model on a richer semantic prefix. 
    \item $a_4$: \textbf{Clarification of Certain Terms} - The query is rewritten to clarify the lexical and semantic meaning of jargons. Since LLMs are pre-trained on broad domain corpora using maximum likelihood next token prediction, rare domain-specific jargons \citep{Clark1983} suffer from sparse exposure and less-calibrated embeddings  \citep{rippeth2023improvingwordsensedisambiguation, peters2018deepcontextualizedwordrepresentations}.

\end{itemize}
{\noindent \bf Contextual Attributes.} 
For each incoming query, we extract a 17-dimensional binary feature vector $\mathbf{f} \in \{0,1\}^{17}$ that captures key linguistic properties as outlined in Table \ref{tab:feature_vector_citations}, grounded by related works in NLP (\S\ref{sec:related}). We have rigorously selected  features that are known to impact both human linguistic understanding and LLM performance. 




{\noindent \bf Reward Model.}
We model each rewritten query’s reward \(r_t\in[0,1]\) as a convex combination of three complementary correctness signals:
\begin{equation}
  r_t \;=\; \alpha\,s_{\mathrm{llm}}
           \;+\;\beta\,s_{\mathrm{fuzz}}
           \;+\;\gamma\,s_{\mathrm{bleu}},
  \qquad
  \alpha+\beta+\gamma = 1,
  \quad
  \alpha,\beta,\gamma\ge0
\end{equation}
\begin{itemize}[noitemsep, leftmargin=*, topsep=0pt, partopsep=0pt, label={\tiny\raisebox{0.5ex}{$\blacktriangleright$}}]
  \item \(s_{\mathrm{llm}}\in\{0,1\}\) is a binary consistency judgment by a GPT-4o-based assessor, calibrated on factual correctness between system and reference answers \citep{liu2023gevalnlgevaluationusing,adlakha2024evaluatingcorrectnessfaithfulnessinstructionfollowing}.  
  \item \(s_{\mathrm{fuzz}}\in[0,1]\) is token‐set similarity from RapidFuzz \citep{max_bachmann_2024_10938887}, capturing soft string overlap.  
  \item \(s_{\mathrm{bleu}}\in[0,1]\) is the BLEU-1 score (unigram precision) under a unit‐cap \citep{Papineni02bleu:a,lin-och-2004-orange,callison-burch-etal-2006-evaluating}, ensuring lexical fidelity.
\end{itemize}
This multi-faceted formulation mitigates individual failure modes inherent in any single metric (e.g. BLEU's paraphrase blindness or edit‐distance oversensitivity) while remaining bounded for stable learning. 
Following \citet{wang-etal-2023-chatgpt}, we leverage the strength of LLMs‐as‐judges, and, as demonstrated by Test-Time RL \citep{zuo2025ttrltesttimereinforcementlearning}, even noisy, self‐supervised signals (e.g.\ pseudo‐labels from majority‐voted LLM outputs) can effectively guide policy updates. 
However, to validate that our convex proxy aligns with human judgment, we assembled a held‐out, manually labeled set of 100 query–answer pairs and measured ROC–AUC of \(r_t\) against binary human labels (Figure~\ref{fig:simplex}). See Alg. \ref{algo:bandits} in Appendix \ref{sec: appendix bandits summary} for further detail.

{\noindent \bf Reward-Weight Simplex Analysis. }
We swept \((\alpha',\beta',\gamma')\) over a triangular grid (\(\alpha'+\beta'+\gamma'=1\)) and computed ROC–AUC on the human‐labeled validation set.  Figure~\ref{fig:simplex} plots each grid point's AUC and overlays the 1\% Pareto frontier (dark region).  Our manual weights \((\alpha,\beta,\gamma)=(0.6,0.3,0.1)\) lie well inside this frontier, demonstrating robustness. The Pareto frontier reveals the following:
\begin{itemize}[noitemsep, leftmargin=*, topsep=0pt, partopsep=0pt, label={\tiny\raisebox{0.5ex}{$\blacktriangleright$}}]
  \item \textbf{LLM‐Judge Robustness (\(\alpha\)):}  
    The ROC–AUC surface is nearly invariant when \(\alpha\) varies by \(\pm0.2\): AUC shifts by <0.5\%, indicating our formulation tolerates large LLM‐judge weight swings.
  \item \textbf{Fuzzy‐Match Sensitivity (\(\beta\)):}  
    Small increases in \(\beta\) rapidly exit the Pareto region, showing that the fuzzy‐match term must be tuned carefully to avoid degrading overall accuracy.
\item \textbf{BLEU‐Only Pitfall (\(\gamma\)):}  
    As \(\gamma\) increases, ROC–AUC steadily declines, bottoming out at \(\gamma=1\) (pure‐BLEU), where the model over‐emphasizes surface overlap at the expense of true correctness.

  \item \textbf{Pareto‐Optimal Region:}  
    Our chosen \((0.6,0.3,0.1)\) sits deep in the high‐AUC plateau, confirming it is a Pareto‐optimal trade‐off among semantic, fuzzy, and lexical signals.
\end{itemize}
Together, these experiments on manually labeled data substantiate our reward design: the LLM‐judge component provides a forgiving anchor, fuzzy‐match demands precise calibration, and BLEU contributes complementary lexical oversight.

\begin{figure}[t]
  \centering
  \begin{subfigure}[t]{0.34\textwidth}
    \includegraphics[width=\textwidth]{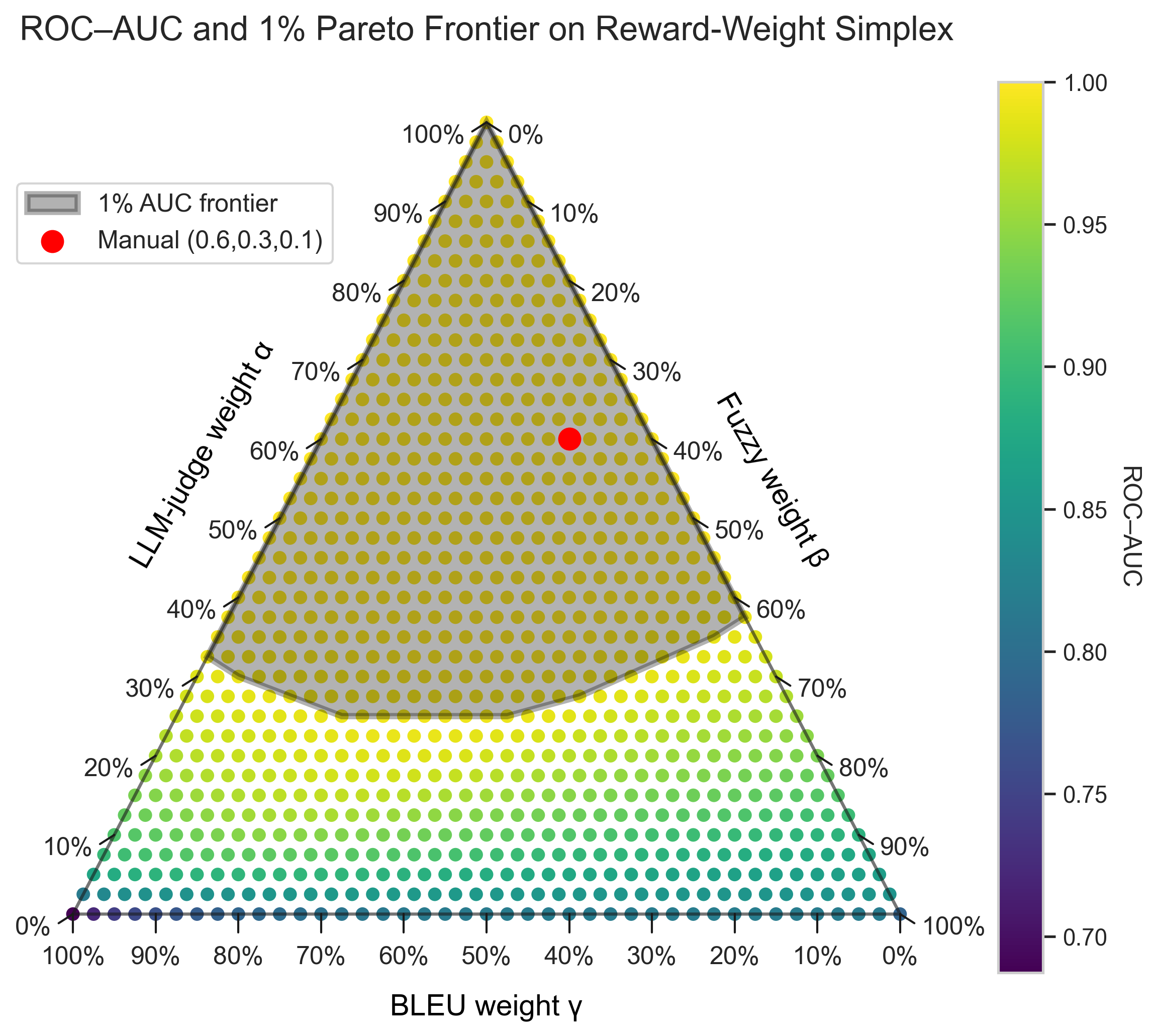}
    \caption{ROC–AUC Pareto frontier on the reward‐weight simplex.}
    \label{fig:simplex}
  \end{subfigure}\hfill
  \begin{subfigure}[t]{0.64\textwidth}
    \includegraphics[width=\textwidth]{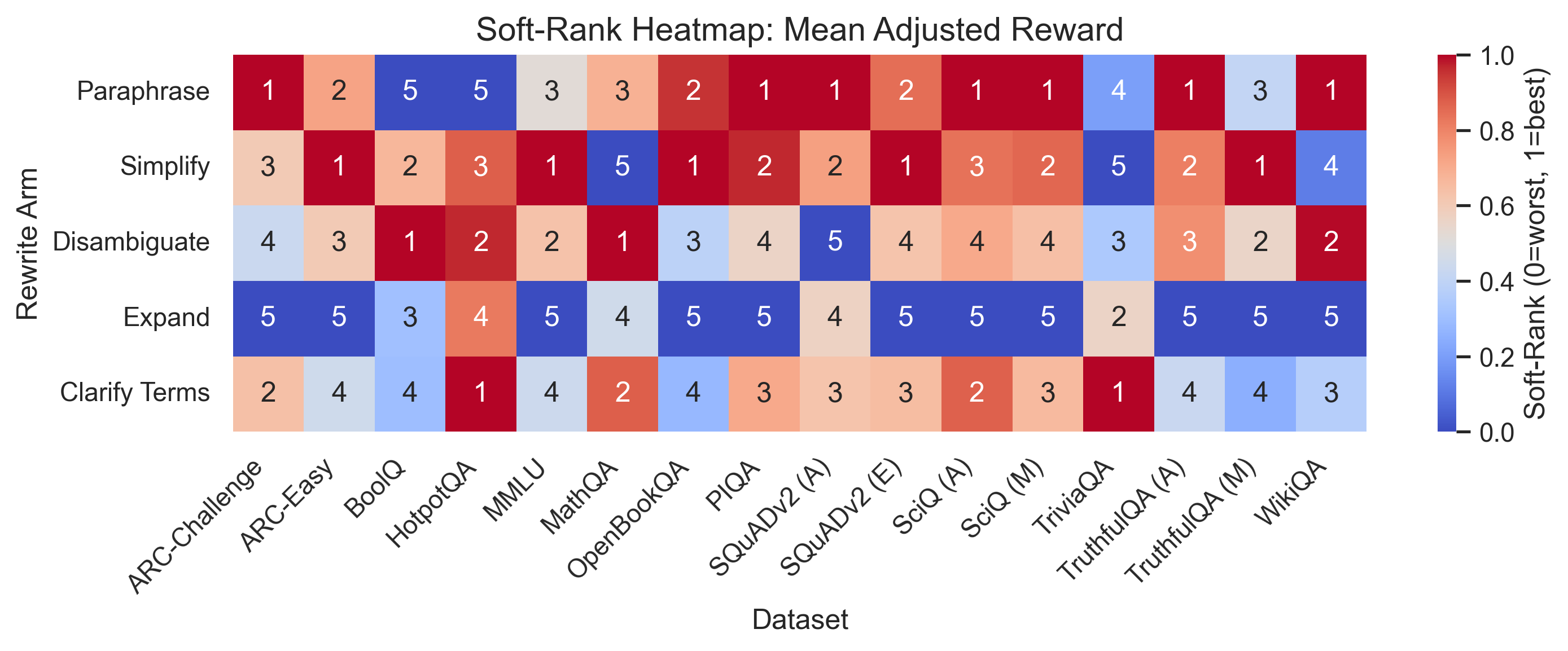}
    \caption{Mean‐reward ranks (1 = best) of each rewrite arm per dataset under our contextual bandit; color intensity indicates closeness to the top rank.}
    \label{fig:rank}
  \end{subfigure}\hfill
  \vspace{-1mm}
  \caption{ 
   (\subref{fig:simplex}) Our chosen $(\alpha,\beta,\gamma)$ lies deep in the 1\% optimal frontier.
    (\subref{fig:rank}) Breakdown of per‐dataset arm performance: different datasets consistently favor different rewrite strategies 
  }
  \label{fig:three_panel}
  \vspace{-5mm}
\end{figure}

\begin{remark} \label{on RL vs. bandits}(On RL Methods vs. Bandits) \ \ \normalfont
Within LLMs, for each input query, the transformer attends over the fixed context window and computes a softmax over the vocabulary to maximize token likelihood \citep{radford2019}.  Consequently, we believe that hallucinations occur at the moment of generation for that single query, making hallucination a \textbf{per-query} phenomenon.  
Indeed, recent PPO variants for LLMs—GRPO \citep{
shao2024deepseekmathpushinglimitsmathematical} and ReMax \citep{li2024remaxsimpleeffectiveefficient}—remove the critic via grouped Monte Carlo or baseline-adjusted returns, highlighting critic-free policies that our bandit formulations naturally generalize.
Therefore, a full-episodic RL problem, which must solve a Markov decision process with long-horizon credit assignment and nonstationary transition dynamics \citep{sutton2018}, can be practically suboptimal. Moreover, many of these methods rely on estimating a fixed average reward or state-action value $\textit{Q(s,a)}$ which can obscure per-query idiosyncrasies; if the optimal rewrite arm varies sharply with linguistic context, a mere empirical average will yield suboptimal policies.
\end{remark}

\begin{remark}(Linkage between Algorithm Choices and RL Methods) \normalfont 
Furthermore, it should be underscored that several algorithms whose usage we investigate here have analogues in RL:  posterior sampling (PSRL) \cite{osband2013more} as an analogue for Thompson sampling \cite{Thompson1933};  follow-the-regularized leader (FTRL) and its variants \cite{shalev2012online}, originate from proximal gradient algorithms \cite{rockafellar1976monotone} whose usage in RL as proximal policy optimization (PPO) \cite{schulman2017proximal} is well-established.
Other PPO-style advances like DAPO \citep{yu2025dapoopensourcellmreinforcement} improve exploration‐exploitation via dynamic sampling and reward filtering, and VAPO \citep{yue2025vapoefficientreliablereinforcement} demonstrates stable Long-CoT training with an explicit value model—showing the spectrum from model-based to model-free approaches that contextual bandits sit within.
%

\end{remark}

{\noindent \bf Choice of Algorithms.}\label{sec:algorithms} 
For \textbf{linear contextual bandits}
we fit a per‐arm linear model \(x_t^\top\theta_k\) and choose either a UCB (LinUCB \citep{Lai1985} / KL‐UCB \citep{garivier2013klucbalgorithmboundedstochastic}), an FTRL regularized weight \citep{mcmahan2015surveyalgorithmsanalysisadaptive}, or a posterior draw (Thompson sampling \citep{Thompson1933}). For \textbf{adversarial bandits}
we consider two parameter‐free adversarial methods—EXP3 \citep{ doi:10.1137/S0097539701398375} and FTPL \citep{KALAI2005291, suggala2020followperturbedleaderoptimism}.  For full scoring, update equations and regret bounds, see Appendix \ref{sec: appendix bandits summary} and Algorithm~\ref{algo:bandits}.

{\noindent \bf Evaluation Metrics.} 
We assess each rewrite policy using three complementary metrics that capture both its ability to explore promising rewrites and its final accuracy relative to a no‐rewrite baseline. All three metrics together give a balanced view of (1) how well a policy explores and exploits, (2) how quickly it converges to good answers, and (3) how often it beats the baseline in reward.


{\noindent \bf Metric 1: Exploration‐Adjusted Reward.}  
Let $r_t\in[0,1]$ be the reward at pull $t$ and let $H_t\in[0,1]$ be the normalized Shannon entropy of the policy’s strategy‐selection history up to $t$. We define the \emph{final exploration‐adjusted reward} as
\[
R_{\mathrm{adj}}
=\sum_{t=1}^T\bigl(r_t+\lambda\,H_t\bigr),
\]
where $\lambda=0.1$ weights the bonus for exploration and $T$ is the trajectory length.  This metric rewards policies that both achieve high per‐pull rewards and maintain sufficient exploration.

{\noindent \bf Metric 2: Mean Cumulative Regret.}  
At each pull we compute instantaneous regret as the gap between the oracle reward (the best achievable rewrite) and the observed reward.  Summing these gives the cumulative regret per run, and we report its average over all runs. Where $r^*_t$ is the maximal reward at pull $t$:
\[
\overline{\text{Regret}}
=\frac1{N}\sum_{\text{runs}}\sum_{t=1}^T\bigl(r^*_t-r_t\bigr),
\]

{\noindent \bf Metric 3: Win Rate vs.\ Baseline. }  
To measure final correctness head‐to‐head, we treat each test pull $t$ as an independent trial and compute the fraction of trials for which a policy's reward $r_t^{\mathrm{policy}}$ strictly exceeds the no‐rewrite baseline's reward $r_t^{\mathrm{base}}$. This directly quantifies how often each rewrite outperforms doing nothing. For $N=100$ test queries, the win rate is
\[
\mathrm{WinRate}
=\frac{1}{N}\sum_{t=1}^N \mathbf{1}\bigl[r_t^{\mathrm{policy}} > r_t^{\mathrm{base}}\bigr] \times 100\%
\]

\section{Experiments and Results}
\label{sec: Experiments}

{\noindent \bf Pipeline.}
For each decision round \(t\):
\[
x_t
\xrightarrow{\;\substack{\text{Extr.\ feat.}\\\mathbf f_t\in\{0,1\}^d}\;}
\mathbf f_t
\xrightarrow{\;\substack{\text{Select}\;a_t\\(\text{rewrite strat.})}\;}
x'_t = g_{a_t}(x_t)
\xrightarrow{\;\mathrm{LLM}\;}
y_t
\xrightarrow{\;\substack{\text{Eval.}\\r_t\in[0,1]}\;}
r_t
\quad
\overset{\text{Update Bandit}}{\looparrowleft}
\]
\begin{enumerate}[noitemsep, leftmargin=*, topsep=0pt, partopsep=0pt]
  \item \textbf{Feature Extraction.} For query \(x_t\), compute \(d\)-dimensional linguistic feature vector \(\mathbf{f}_t \in \{0,1\}^d\).
  \item \textbf{Arm Selection.} The bandit receives \(\mathbf{f}_t\) and selects a rewrite arm \(a_t \in \{1,\ldots,K\}\).
  \item \textbf{Query Rewriting.} Apply the selected arm to obtain the candidate query \(
    x'_t \;=\; g_{a_t}(x_t)\,.
  \)
  \item \textbf{LLM Inference.} Issue \(x'_t\) to \texttt{gpt-4o-2024-08-06}, producing response \(y_t\).
  \item \textbf{Reward Evaluation.} Compute scalar reward \(r_t \in [0,1]\) via our reward formulation.
  \item \textbf{Bandit Update.} Update the internal state of the bandit  based on \((a_t, r_t)\).
\end{enumerate}

{\noindent \bf Dataset and Query Construction.}
We evaluate on \(D=13\) diverse QA benchmarks (see Table~\ref{tab:dataset_details}).  For each dataset, we sample \(\lvert \mathcal{Q}\rvert\) queries satisfying:
(1) Original Answerability: the query in the dataset (\textit{q}) is answered correctly by \texttt{gpt-4o-2024-08-06}, and
(2) Perturbation Validity: of its five lexically perturbed but semantically invariant versions of the dataset's query, measured by numerous metrics such as LLM-as-judge and n-gram based metrics \citep{lin-2004-rouge, Papineni02bleu:a, wang-etal-2023-chatgpt, fu2023gptscoreevaluatedesire}, between one and three perturbations yield incorrect answers.
Then, we randomly choose \(x_t\) in $\lvert \mathcal{Q}\rvert\ $ to train QueryBandits.


The importance of this query construction process deserves emphasis. Through our investigations, we have discovered that the ubiquity of benchmark datasets in Table~\ref{tab:dataset_details} within pre-training and fine-tuning regimes has engendered a potentially pernicious form of prompt memorization.  When we deploy our linear contextual bandits, the policy converges almost exclusively to no-rewriting, effectively demonstrating that the LLM has most likely memorized these exact phrasings rather than learning to exploit deeper linguistic structure (Figure~\ref{fig:no-rewrite}). To evaluate genuine rewrite efficacy---and to prevent our results from collapsing into a trivial memorization baseline---we therefore choose a perturbed version of the dataset's query to construct the \textbf{incoming query} for our bandits. This experimental setup thus prioritizes query-rewriting strategies that are non-degenerate.

{\noindent \bf Experimental Configuration.}
We compare three non-contextual and six linear contextual bandits against zero‐shot prompting and a no‐rewrite baseline. All reported metrics are averaged over all dataset runs per algorithm. 
We compare \(M\) bandit algorithms and prompting strategies over \(K=5\) rewrite arms.  Each algorithm runs for \(T=\lvert \mathcal{Q}_{D}\rvert\) rounds on each of the \(D\) datasets (Table~\ref{tab:dataset_details}).  Thus, 
$
  \text{Total Pulls} \;=\; M \;\times\; D \;\times\; \lvert \mathcal{Q}_{D}\rvert\, = 253,440,
$
with \(\lvert \mathcal{Q}_{D}\rvert \approx 1050\), \({M = 15}\), and \({D = 16}\).  We bootstrap samples with replacement for TruthfulQA to obtain approximately 1050 queries. Hyperparameters (learning rates, exploration coefficients, regularization constants) are tuned via grid search on a held-out validation set.

\begin{figure}[t]
  \centering
  \includegraphics[width=\linewidth]{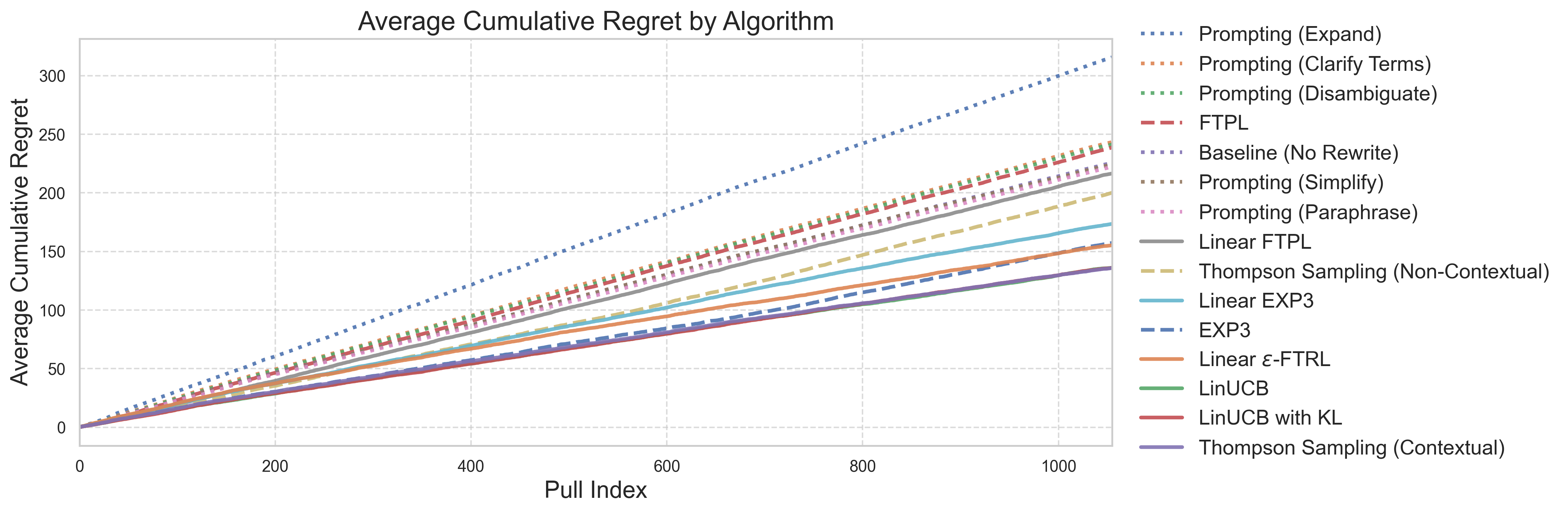}
  \caption{Cumulative reward {\it averaged across all datasets} over the no‐rewrite baseline for each algorithm (sorted by final performance), highlighting the superior gains achieved by contextual bandits compared to non‐contextual learners and static prompt‐based rewrites. 
  }
  \label{fig:excess_regret}
  \vspace{-5mm}
\end{figure}

\begin{table}[t]
\centering
\small
\caption{Rewrite‐policy performance: final cumulative exploration‐adjusted reward, mean cumulative regret, and win rate vs.\ no‐rewrite baseline.} 
\label{tab:bandit_results}
\resizebox{0.9\linewidth}{!}{
\begin{tabular}{l c r r r}
\toprule
\textbf{Algorithm}                      & \textbf{Contextual?} & \textbf{Adj.\ Reward} \(\uparrow\) & \textbf{Cum.\ Regret} \(\downarrow\) & \textbf{Win Rate} \(\uparrow\) \\
\midrule
Thompson Sampling (Contextual)         & \ding{51}   & 819.04 & 135.84 & 87.5\% \\
LinUCB with KL                         & \ding{51}   & 818.79 & 136.00 & 87.0\% \\
LinUCB                                 & \ding{51}   & 818.60 & 136.12 & 86.9\% \\
Linear \(\epsilon\)-FTRL               & \ding{51}   & 799.57 & 155.30 & 85.0\% \\
EXP3 (Non-Contextual)                  & \ding{55}   & 797.47 & 157.31 & 86.5\% \\
Linear EXP3                            & \ding{51}   & 781.05 & 173.60 & 83.8\% \\
Thompson Sampling (Non-Contextual)     & \ding{55}   & 754.66 & 200.18 & 81.7\% \\
Linear FTPL                            & \ding{51}   & 738.07 & 216.54 & 76.3\% \\
FTPL (Non-Contextual)                  & \ding{55}   & 716.05 & 238.85 & 62.8\% \\
\midrule
Prompting (Paraphrase)               & –   & 732.39 & 222.56 & 44.9\% \\
Prompting (Simplify)                 & –   & 730.13 & 224.42 & 50.1\% \\
Prompting (Disambiguate)             & –   & 713.65 & 241.25 & 42.4\% \\
Prompting (Clarify Terms)            & –   & 711.65 & 243.35 & 38.2\% \\
Prompting (Expand)                   & –   & 639.25 & 315.71 & 27.2\% \\
\midrule
Baseline (No Rewrite)                  & –             & 729.20 & 225.85 & –      \\
\bottomrule
\end{tabular}
}
\vspace{-2mm}
\end{table}


{\noindent \bf Hypothesis 1: Can QueryBandits reduce hallucination?}
Table~\ref{tab:bandit_results} and Figure~\ref{fig:excess_regret} compares our QueryBandit algorithms against the baseline and five static prompting scenarios on 13 QA benchmarks (1,050  queries per dataset). Our top contextual learner—Thompson Sampling with the 17-dimensional feature vector—achieves an \textbf{87.5\% win rate} and 819.04 exploration-adjusted reward, compared to the \textsc{Baseline: No Rewrite}, signifying that contextual query rewriting can reduce hallucination. As a side note: we aggregate results across datasets rather than report per-dataset Monte Carlo trials, as within-dataset permutation yielded trivial randomization.

{\noindent \bf Hypothesis 2: Can QueryBandits outperform prompting?}
Our best performing bandit, Thompson Sampling,   exceeds the performance of static prompting 
for \textsc{Prompting (Paraphrase)} (44.9\% / 732.39) and \textsc{Prompting (Expand)} (27.2\% / 639.25), as seen in Table~\ref{tab:bandit_results} and Figure \ref{fig:excess_regret}. These gains confirm that raw static prompting cannot approach the effectiveness of a learner that adapts its rewrite choice to each query’s linguistic fingerprint. By framing rewrite selection as an online decision problem and leveraging per-query context, QueryBandits allocate exploration where uncertainty is high and exploitation where features reliably predict hallucination risk—resulting in up to double the hallucination reduction of any static strategy, with no additional model fine-tuning. 

{\noindent \bf Hypothesis 3: Do linear contextual bandits outperform those which are oblivious to context?}
Crucially, ablating the 17-dimensional feature input causes Thompson Sampling’s performance to drop to just 81.7\% win rate and 754.66 exploration-adjusted reward (–5.8 pts, –64.38 reward points), despite identical hyperparameters and run count. Since our reward directly measures output correctness, this performance gap confirms that the linguistic features carry associative signal about hallucination risk. Moreover, the majority of our linear QueryBandits outperform those which are oblivious to context (Fig.~\ref{fig:excess_regret}), signifying the importance of taking context into account in rewrites.

\begin{figure}[t] 
   \begin{minipage}[t]{0.48\textwidth}
   \centering
  \includegraphics[width=\textwidth]{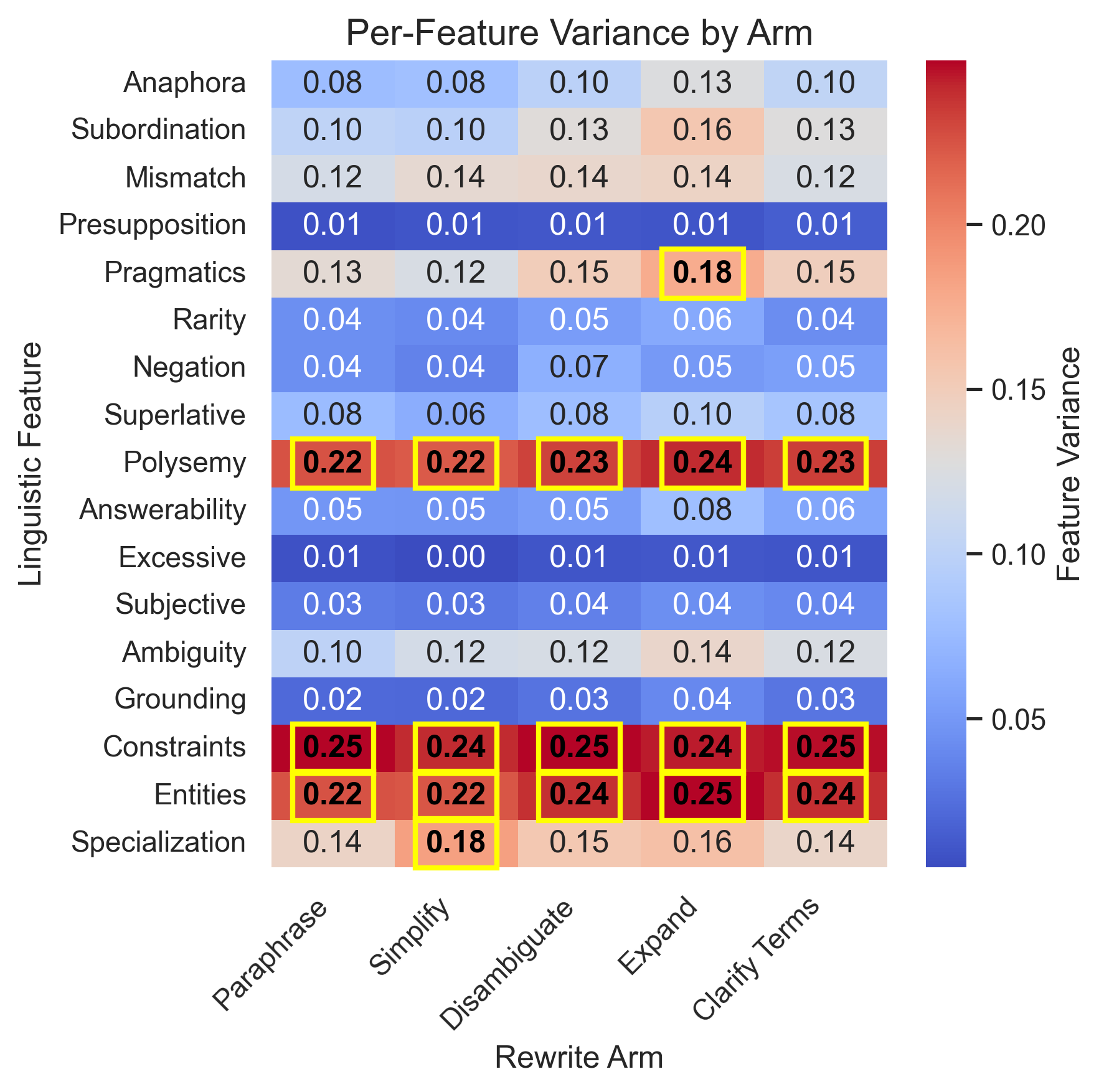}
  \caption{%
    \textbf{Contextual Per‐Feature Variance by Arm.} For each arm, we compute the variance of each binary linguistic feature over all queries on which that arm was chosen. High variance means the bandit frequently switches the arm on that feature’s presence. 
  }
  \label{fig:rel-feature-contrib}
  \end{minipage}%
  \hfill
   \begin{minipage}[t]{0.48\textwidth}
   \centering
  \includegraphics[width=\textwidth]{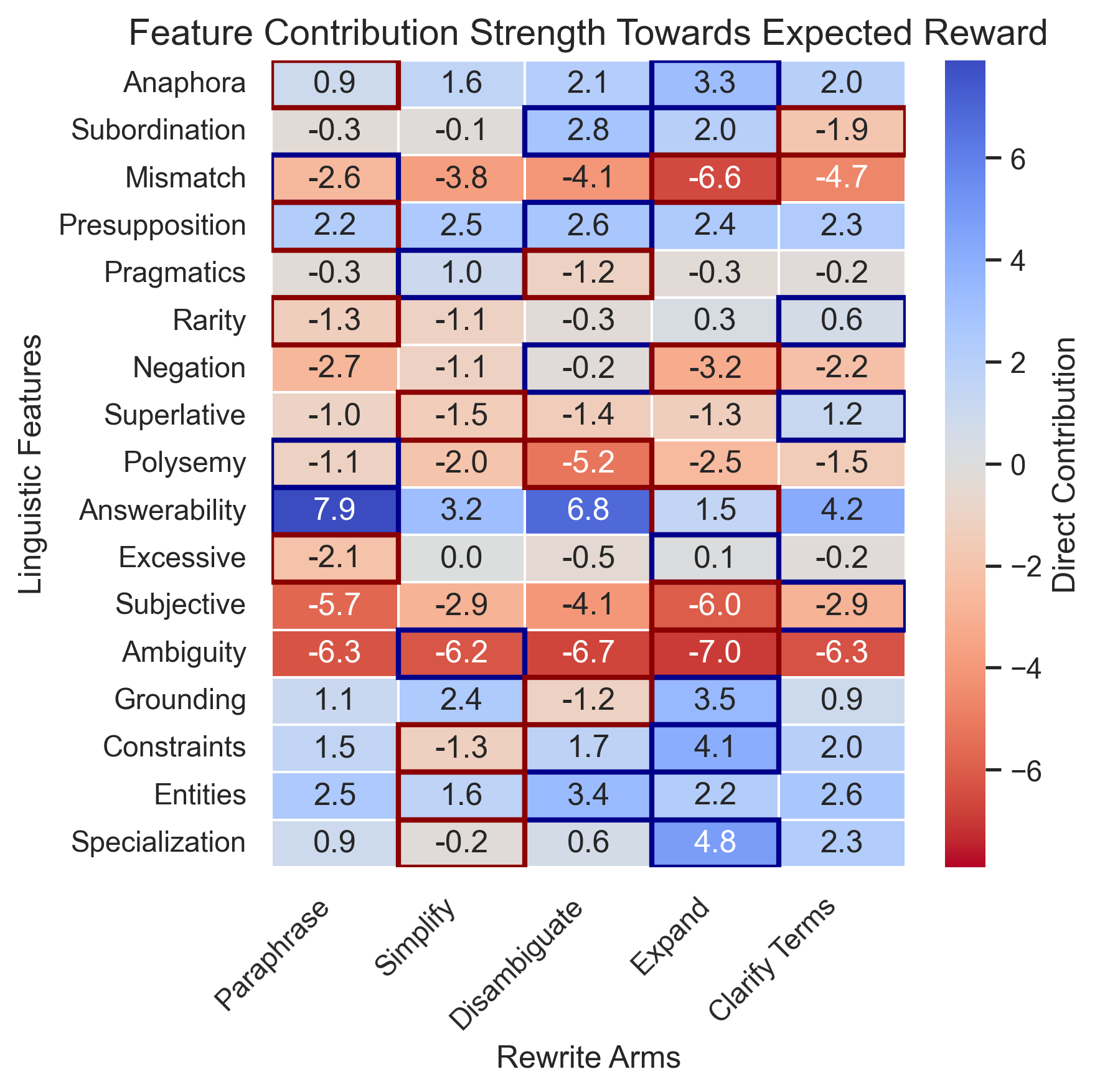}
  \caption{%
    \textbf{Contextual Feature Contribution Strength.} These are the averaged $\theta$ weights (direct contributions) of each feature to the expected reward under each arm. Positive weights indicate features that boost that arm’s reward; negative weights indicate features that \textit{penalize} it. 
  }
  \label{fig:raw-feature-contrib}
  \end{minipage}%
  \vspace{-5mm}
\end{figure}

{\noindent \bf Hypothesis 4: Is there an association between query features and reward?} 
Each rewrite arm seems to exhibit different sensitivities toward different \textit{linguistic features}. 
Figures~\ref{fig:rel-feature-contrib} \& \ref{fig:raw-feature-contrib} plot each arm’s average variance and learned regression weights $\theta$ across 17 binary linguistic features. Interestingly, the same linguistic feature can flip from strongly sensitive for one arm to insensitive for another---for example, the feature \emph{(Domain) Specialization} is quite sensitive to the arm/strategy \textsc{Expand} but relatively much less to \textsc{Simplify}. A plausible explanation might be that domain specific queries inherently require specialized context outside the LLM’s broad‐domain priors, so \textsc{Expand}—by injecting explicit entity attributes or ontological qualifiers—reinforces the model’s semantic grounding, whereas \textsc{Simplify} risks excising those critical signals. While our association matrix is interesting, each learned coefficient 
does not strictly prove a causal mechanism. 

{\noindent \bf Hypothesis 5: Is there a \textit{single} rewrite strategy that maximizes reward for all types of queries?}
Our analysis of the per-arm regression weights (Figure~\ref{fig:raw-feature-contrib}) reveals that no single rewrite strategy dominates across all linguistic profiles.  Instead, each arm’s effectiveness hinges on a distinct “feature footprint”. 
For example, \textsc{Simplify} thrives when pragmatic cues (e.g.\ discourse markers, politeness markers) are present—these guide safe syntactic pruning—but falters on superlative constructions, whose removal strips away essential comparative meaning. For our interpretation of these sharp inversions  feature–arm interactions, refer to Appendix Table~\ref{tab:arm-effects}. Therefore, our results demonstrate that the diversity of arm selected is correlated with feature variance---and that there is no \textit{single} rewrite arm that fits all queries.

\section{Conclusion}
\label{sec: conclusion}

We have presented a novel bandit framework of query rewriting to mitigate hallucinations. By comparing QueryBandits against a \textsc{no-rewrite} baseline and static prompting strategies across diverse benchmarks, our best-performing contextual bandit consistently outperforms by a considerable margin in terms of win-rates and cumulative regret. We have also learnt that rewrite strategies are sensitive to linguistic features, and that we can exploit these features for no-regret rewriting. Through a purely forward-pass mechanism, QueryBandits delivers meaningful results in terms of not only hallucination mitigation but also paves the way for LLM interpretability. Future work can dive deeper into casual inference techniques and capture pairwise polynomial feature interactions within the query---and thus, further enhance trustworthiness of LLM systems and their societal value (\ref{sec: extended limitations}).

\section*{Disclaimer}
{
This paper was prepared for informational purposes by the Artificial Intelligence Research group of JPMorgan Chase \& Co. and its affiliates ("JPMorgan'') and is not a product of the Research Department of JPMorgan. JPMorgan makes no representation and warranty whatsoever and disclaims all liability, for the completeness, accuracy or reliability of the information contained herein. This document is not intended as investment research or investment advice, or a recommendation, offer or solicitation for the purchase or sale of any security, financial instrument, financial product or service, or to be used in any way for evaluating the merits of participating in any transaction, and shall not constitute a solicitation under any jurisdiction or to any person, if such solicitation under such jurisdiction or to such person would be unlawful.
}

{
\small
\bibliographystyle{plainnat}
\bibliography{custom}
}


\appendix

\section{Appendix / supplemental material}

\subsection{Limitations}
\label{sec: extended limitations}
Current limitations in our work are as follows: our current contextual bandit framework treats each of the 17 features as independent, but does not capture higher-order interactions. This can provide an exciting avenue of future research in terms of measuring whether the combination of features jointly exacerbates hallucination. Likewise, we would like to highlight that the feature-arm regression weights do not stipulate a causal relationship - highly sophisticated causal relationships are difficult to formulate within LLMs due to the inherent difficulties of interpreting a neural network's internal layers; thus, in this paper, we focus on providing empirical studies and the conclusions we can draw from them. Finally, even with our rigorous studies to find the ROC-AUC Pareto-frontier, our reward model leverages LLM-as-judge, which may reflect the LLM's bias. Overall, these limitations posit potential directions by which the research community can further pursue - and ultimately help expand our understanding of these powerful, albeit hallucinatory models.

\begin{figure*}[t]
  \centering
  \begin{minipage}[t]{0.49\textwidth}
    \centering
    \includegraphics[width=\textwidth]{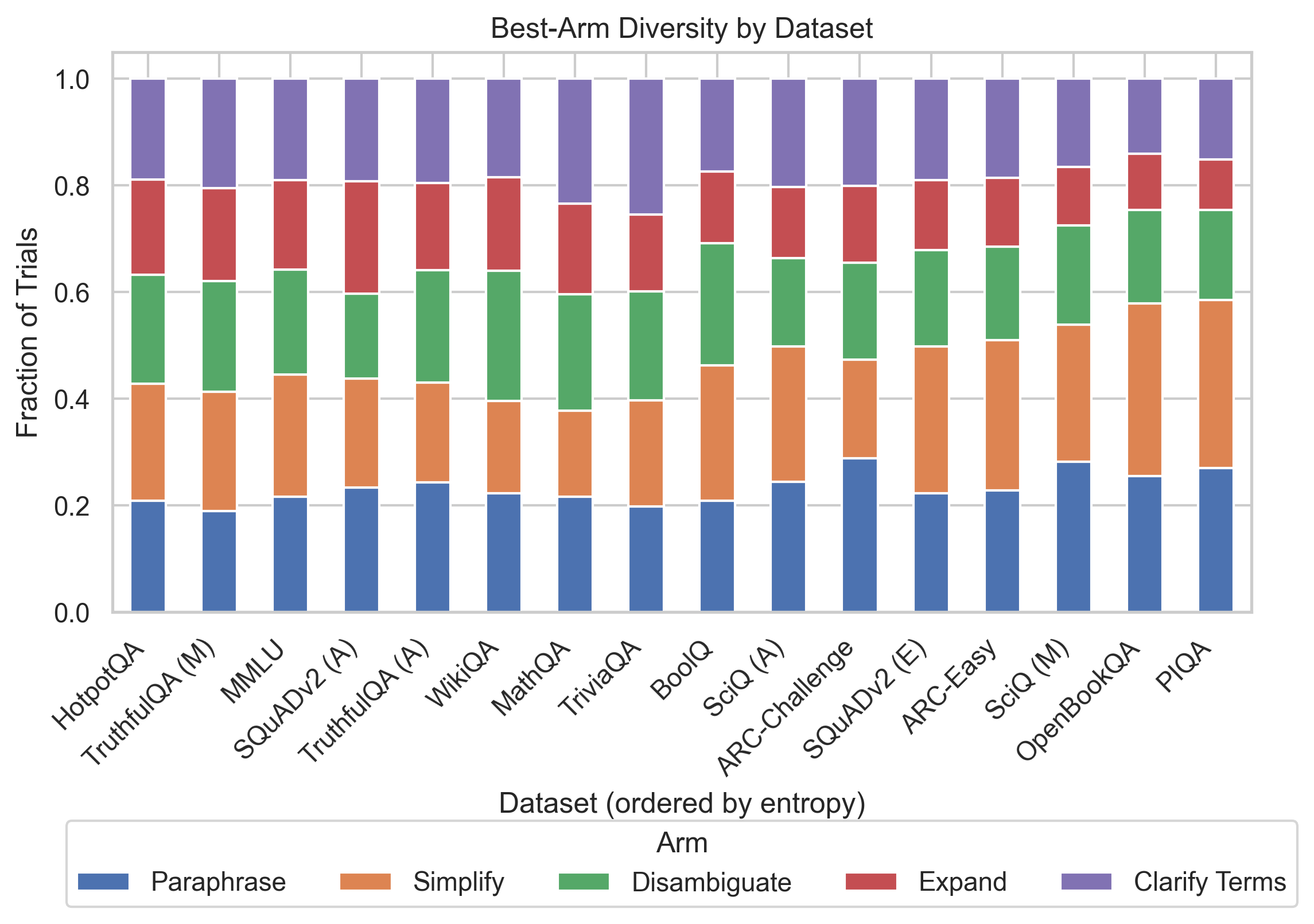}
    \subcaption{Arm Diversity for Contextual Bandits, as a Fraction of Trials.}
    \label{fig:lin_diversity}
  \end{minipage}\hfill
  \begin{minipage}[t]{0.49\textwidth}
    \centering
    \includegraphics[width=\textwidth]{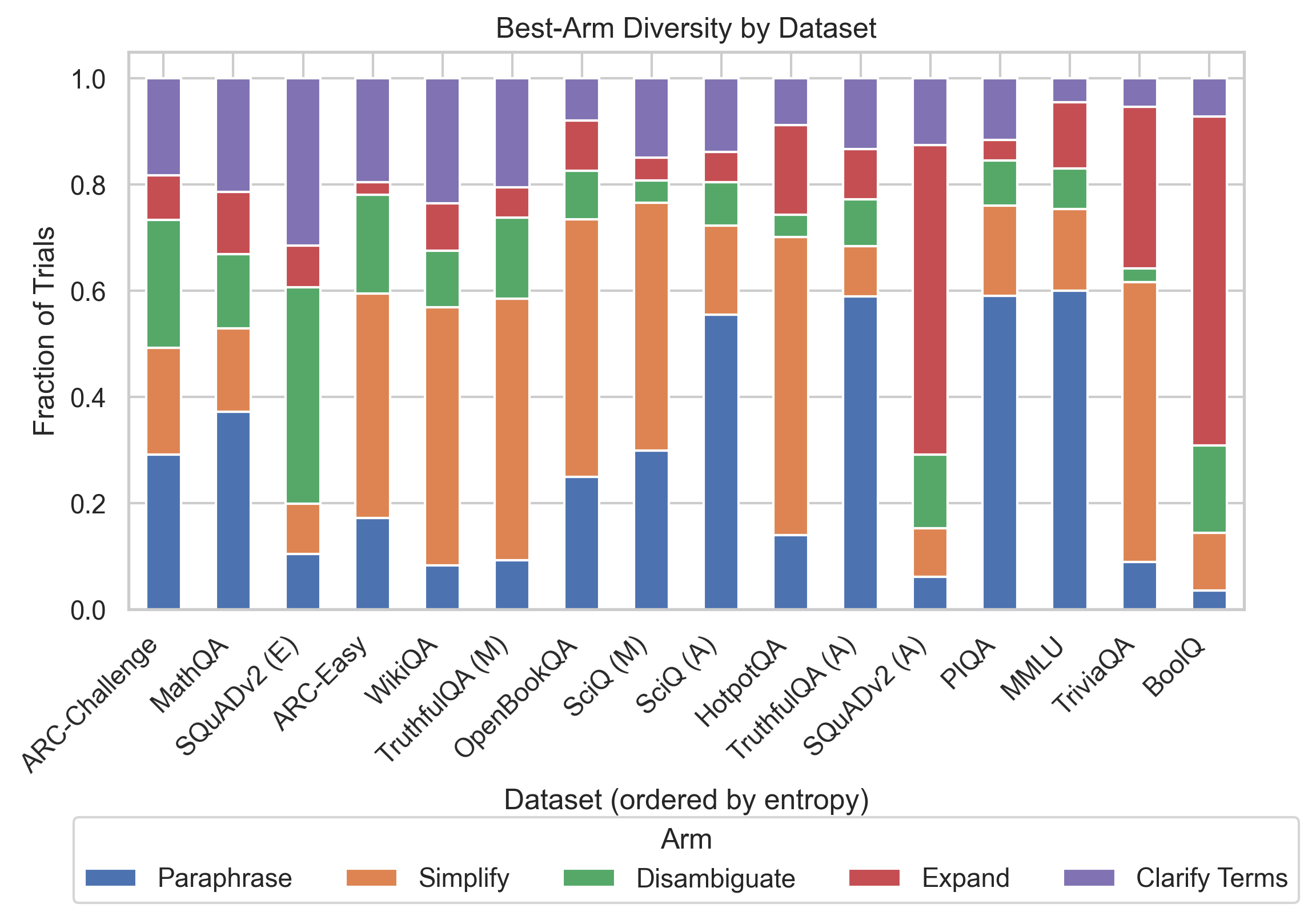}
    \subcaption{Arm Diversity for Non-Contextual Bandits, as a Fraction of Trials.}
    \label{fig:non_lin_diversity}
  \end{minipage}
  \caption{For Non-Contextual bandits, \textit{almost every} dataset is dominated by a single arm with the highest global reward (typically 40\%-60\% of the trials). The remaining 40-60\% is split among the other four arms as noise, the non-contextual policy has no way to "know" when within a dataset a different arm might do better. In contrast, Contextual bandits show a more even mix: the top arm is only $\sim$25-30\%, with two or three other arms contributing sizable shares (15-25\% each). The contextual policy \textit{reads the features} and diversifies its choices within each dataset.}
  \label{fig:linvnonlin_diversity}
\end{figure*}

\subsection{Summary of Bandits}
\label{sec: appendix bandits summary}
\begin{itemize}[noitemsep, leftmargin=*, topsep=0pt, partopsep=0pt, label={\tiny\raisebox{0.5ex}{$\blacktriangleright$}}]
  \item \textbf{Non‐Contextual Adversarial}
    \begin{itemize}[noitemsep,leftmargin=*, topsep=0pt, partopsep=0pt]
      \item \textbf{EXP3} \citep{ doi:10.1137/S0097539701398375}  
        Maintains weights \(w_k\), samples \(a_t\propto w_k\), updates  
        \(\;w_{a_t}\leftarrow w_{a_t}\exp\bigl(\tfrac{\gamma\,r_t}{K\,p_{a_t}}\bigr).\)
      \item \textbf{FTPL} \citep{KALAI2005291, suggala2020followperturbedleaderoptimism}  
        Adds Gumbel noise \(\xi_k\!\sim\!\mathrm{Gumbel}(0,1/\eta)\) \citep{Gumbel1941} to cumulative rewards,  
        selects \(a_t=\arg\max(\mathrm{cum\_reward}_k+\xi_k)\), then increments the chosen arm’s reward.
    \end{itemize}

  \item \textbf{Contextual Stochastic}
    \begin{itemize}[noitemsep,leftmargin=*, topsep=0pt, partopsep=0pt,]
      \item \textbf{LinUCB} \citep{Lai1985}  
        Selects  
        \(\;a_t=\underset{k}{\arg\max}\bigl(x_t^\top\hat\theta_k + \alpha\sqrt{x_t^\top A_k^{-1}x_t}\bigr)\),  
        updates \(A_k\!\leftarrow\!A_k + x_t x_t^\top,\;b_k\!\leftarrow\!b_k + r_t x_t\).
      \item \textbf{KL‐UCB (LinUCB-KL)} \citep{garivier2013klucbalgorithmboundedstochastic}  
        Replaces the UCB term with a KL-divergence‐based confidence bound.
      \item \textbf{Thompson Sampling} 
        Maintains Gaussian posterior \(\mathcal{N}(\mu_k,\Sigma_k)\); samples  
        \(\tilde\theta_k\), picks \(a_t=\arg\max x_t^\top\tilde\theta_k\), updates the posterior.
    \end{itemize}

  \item \textbf{Contextual Adversarial}
    \begin{itemize}[noitemsep,leftmargin=*, topsep=0pt, partopsep=0pt,]
      \item \textbf{FTRL} \citep{mcmahan2015surveyalgorithmsanalysisadaptive}  
        Selects arm maximizing \(x_t^\top w_k - \lambda\|w_k\|_1\), with an \(\ell_1\) regularizer.
     \item \textbf{$\epsilon$-greedy FTRL} ...
      \item \textbf{LinearEXP3} \citep{pmlr-v125-neu20b}  
        Contextual extension of EXP3, sampling arms based on exponentiated linear scores.
      \item \textbf{LinearFTPL} \citep{Hanna1957}  
        Contextual adaptation of FTPL, applying Gumbel perturbations to linear reward estimates.
    \end{itemize}
\end{itemize}

\begin{algorithm}[t]
\caption{General Bandit + Rewrite Loop}
\label{algo:bandits}
\begin{algorithmic}[1]
  \Require arms \(\mathcal{A}\), context \(x_t\), algorithm \(\mathrm{algo}\in\{\text{EXP3, FTPL, LinUCB, KL, FTRL, Thompson}\}\), hyperparameters
  \For{\(t = 1\) to \(T\)}
    \State observe \(x_t\)
    \For{each arm \(k\in\mathcal A\)}
      \State \(s_k \leftarrow \mathrm{Score}(\mathrm{algo}, k, x_t)\)
    \EndFor
    \State select \(a_t = \arg\max_{k\in\mathcal A} s_k\)
    \State apply rewrite \(a_t\) to query and observe reward \(r_t\)
    \State \(\mathrm{Update}(\mathrm{algo}, a_t, x_t, r_t)\)
  \EndFor
\end{algorithmic}
\end{algorithm}

\subsection{LinUCB} 
The estimated parameter is:
\begin{equation}
    \hat{\theta}_a = A_a^{-1} \mathbf{b}_a.
\end{equation}
Given a query feature vector $\mathbf{x}$, the upper confidence bound (UCB) for arm $a$ is:
\begin{equation}
\text{UCB}_a(\mathbf{x}) = \mathbf{x}^\top \hat{\theta}_a + \alpha \sqrt{\mathbf{x}^\top A_a^{-1} \mathbf{x}},
\end{equation}
where $\alpha$ controls the exploration–exploitation trade-off. The arm selected is:
\begin{equation}
a^* = \argmax_{a \in \mathcal{A}} \text{UCB}_a(\mathbf{x}).
\end{equation}
Upon observing reward $r$, update:
\begin{equation}
A_a \leftarrow A_a + \mathbf{x}\mathbf{x}^\top,\quad \mathbf{b}_a \leftarrow \mathbf{b}_a + r\,\mathbf{x}.
\end{equation}

\begin{figure*}[t]
  \centering
  \begin{minipage}[t]{0.59\textwidth}
    \centering
    \includegraphics[width=\textwidth]{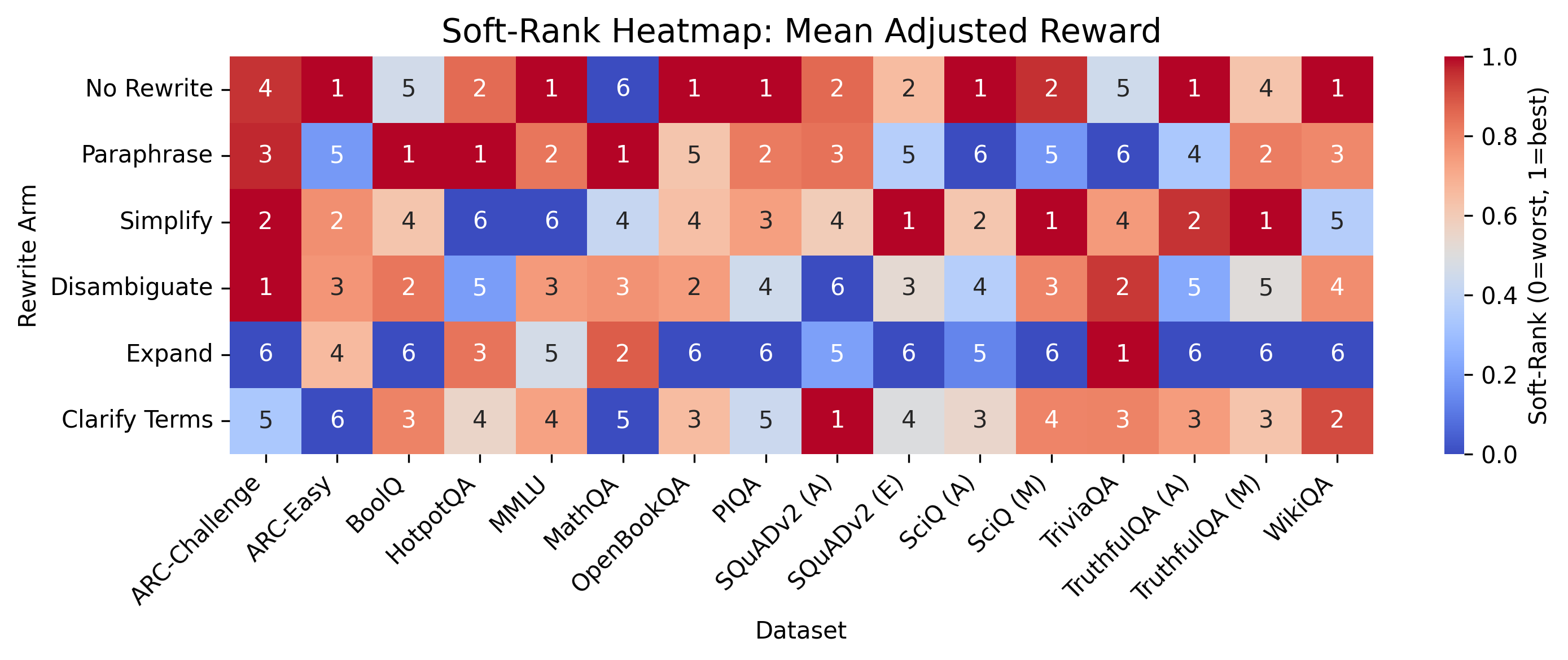}
    \subcaption{Soft Rank Heatmap for all Bandits, including arm \textsc{No Rewrite}.}
    \label{fig:no-rewrite-heatmap}
  \end{minipage}\hfill
  \begin{minipage}[t]{0.38\textwidth}
    \centering
    \includegraphics[width=\textwidth]{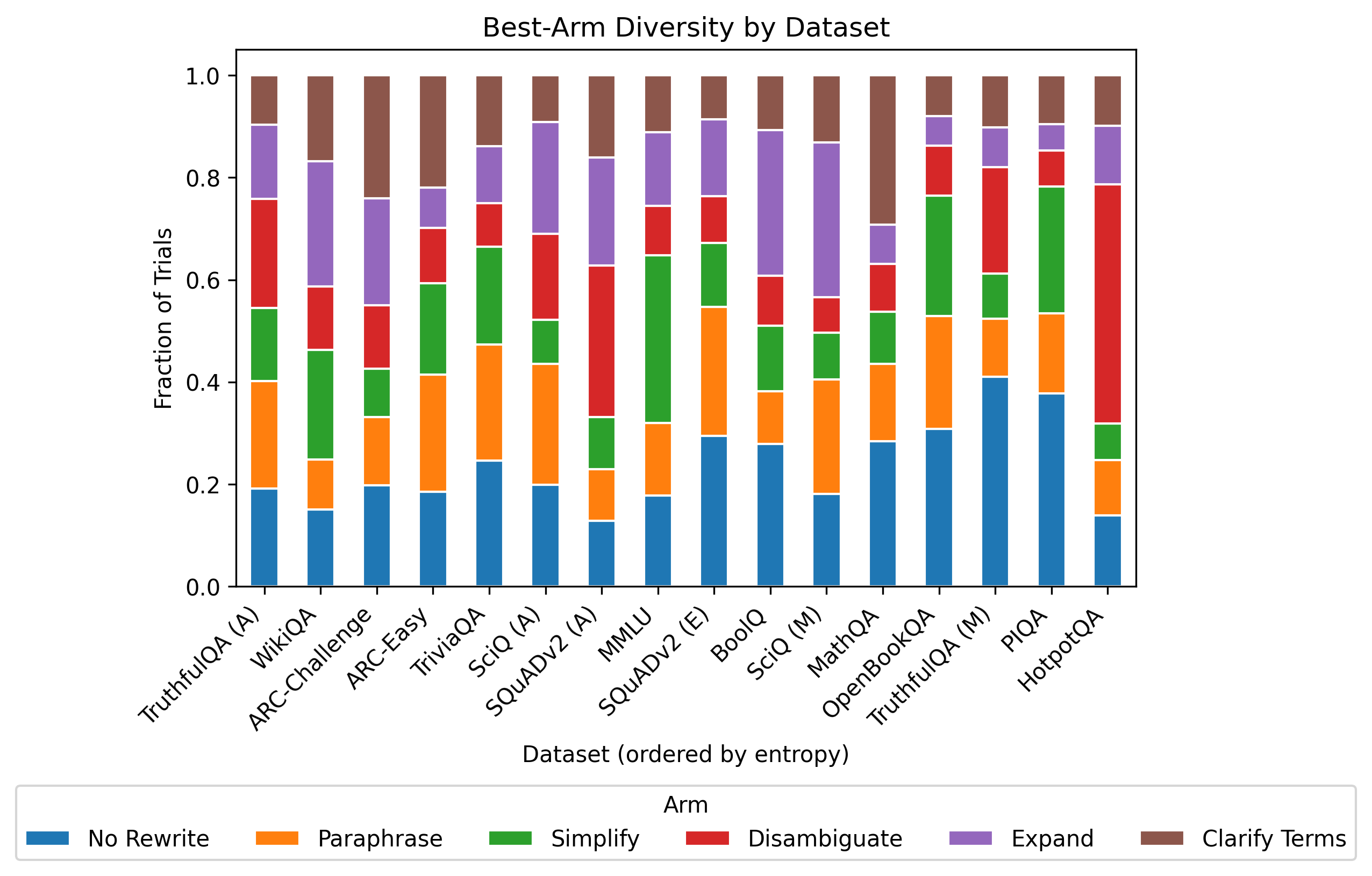}
        \subcaption{Arm Diversity when including \textsc{No Rewrite}.}
    \label{fig:no-rewrite-fraction}
  \end{minipage}
  \caption{\textbf{Impact of the No‐Rewrite Arm.}  
Note that these experiments are conducted on the original query "as-is" in the benchmark dataset, with no perturbations. Upon enabling the \textsc{No Rewrite} option, our contextual bandit rapidly converges to this arm, which then achieves the highest reward on several datasets. We attribute this behavior to the LLM’s tendency to memorize benchmark questions. }
  \label{fig:no-rewrite}
\end{figure*}

\subsection{\textbf{LinUCBKL Bandit Strategy:}}
The algorithm is initialized with parameters: number of arms $n_{\text{arms}}$, dimension $d$, regularization parameter $\lambda$, exploration parameter $\alpha$, noise variance $\sigma_{\text{noise}}$, and KL-bound constant $c$. Each arm $a$ maintains a matrix $\mathbf{A}_a$ and a vector $\mathbf{b}_a$, initialized as $\lambda \mathbf{I}_d$ and $\mathbf{0}_d$, respectively.

The \texttt{select\_arm} method computes the score for each arm $a$ using the following formulation:
\begin{align*}
\theta_a &= \mathbf{A}_a^{-1} \mathbf{b}_a \\
\mu_a &= \mathbf{x}^\top \theta_a \\
\text{var}_a &= \mathbf{x}^\top \mathbf{A}_a^{-1} \mathbf{x} \\
n_a &= \max(1, \text{counts}[a]) \\
\text{raw\_bound}_a &= \frac{\log(t) + c \log(\log(t+1))}{n_a} \\
\text{bound}_a &= \max(\text{raw\_bound}_a, 0.0) \\
\text{bonus}_a &= \sqrt{2 \cdot \text{var}_a \cdot \text{bound}_a} \\
\text{score}_a &= \mu_a + \text{bonus}_a
\end{align*}
where $\mathbf{x}$ is the context vector, $t$ is the time step, and $\text{counts}[a]$ is the number of times arm $a$ has been selected. The arm with the highest score is selected for exploration.

The \texttt{update} method updates the matrix $\mathbf{A}_a$ and vector $\mathbf{b}_a$ for the selected arm $a$ based on the received reward $r_t$:
\begin{align*}
\mathbf{A}_a &\leftarrow \mathbf{A}_a + \mathbf{x} \mathbf{x}^\top \\
\mathbf{b}_a &\leftarrow \mathbf{b}_a + r_t \mathbf{x} \\
\text{counts}[a] &\leftarrow \text{counts}[a] + 1
\end{align*}

This strategy leverages the KL-bound to dynamically adjust exploration bonuses, enhancing the LinUCB algorithm's ability to balance exploration and exploitation in a contextual setting.

\begin{figure*}[t]
  \centering
  \begin{minipage}[t]{0.49\textwidth}
    \centering
    \includegraphics[width=\textwidth]{feature_var_linear.png}
    \subcaption{Contextual Model Feature Variance.}
    \label{fig:feat_var_lin}
  \end{minipage}\hfill
  \begin{minipage}[t]{0.49\textwidth}
    \centering
    \includegraphics[width=\textwidth]{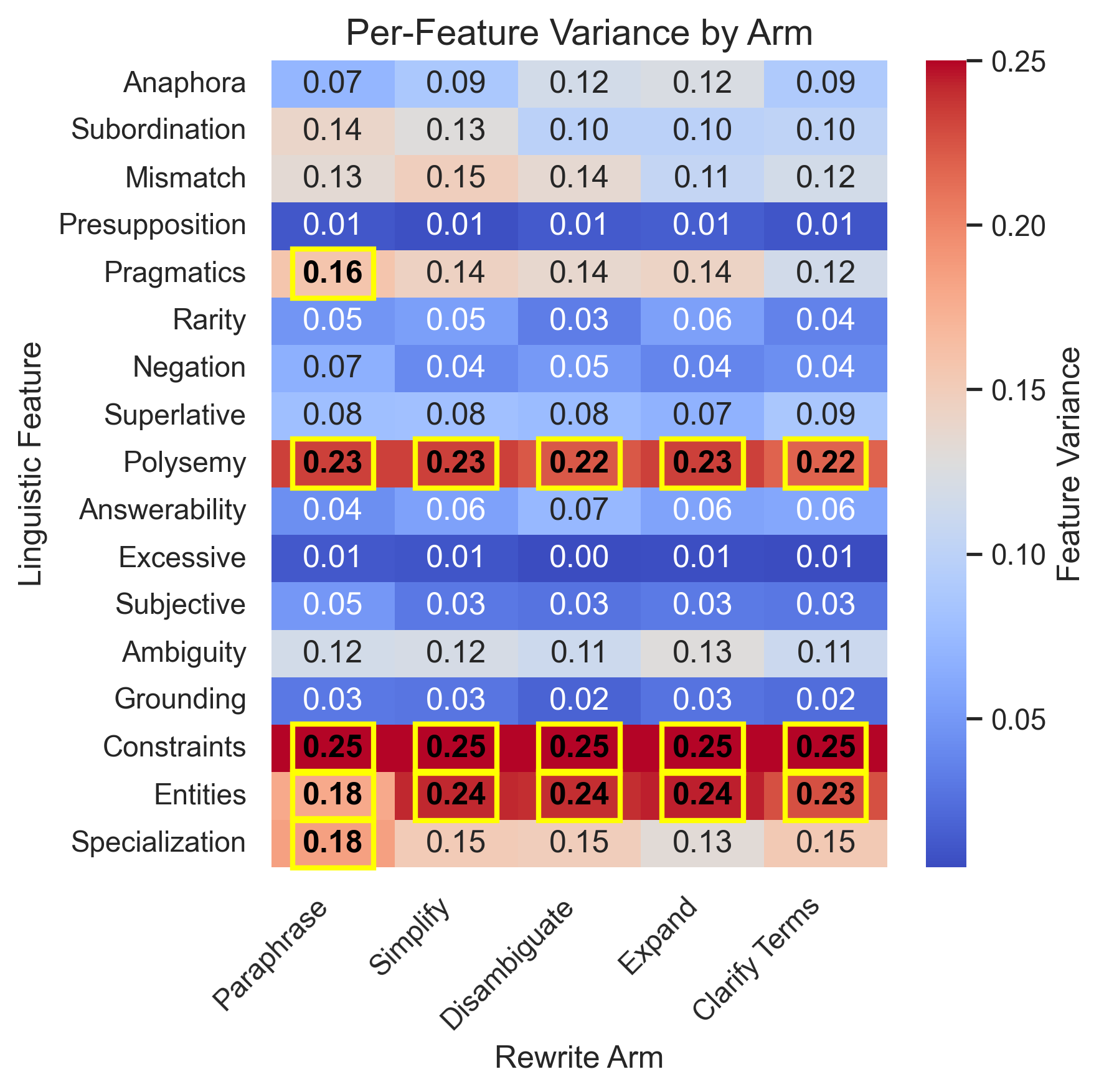}
        \subcaption{Non-Contextual Model Feature Variance.}
    \label{fig:feat_var_nonlin}
  \end{minipage}
  \caption{Comparison of Feature Variance between (\subref{fig:feat_var_lin}) our contextual bandits and (\subref{fig:feat_var_nonlin}) its non-contextual counterparts. \textit{Polysemy}, \textit{Constraints} and \textit{Entities} show the most variation. \textit{Presupposition}, \textit{Excessive Details}, and \textit{Grounding} have the least.}
  \label{fig:var}
\end{figure*}

\begin{figure*}[t]
  \centering
  \begin{minipage}[t]{0.49\textwidth}
    \centering
    \includegraphics[width=\textwidth]{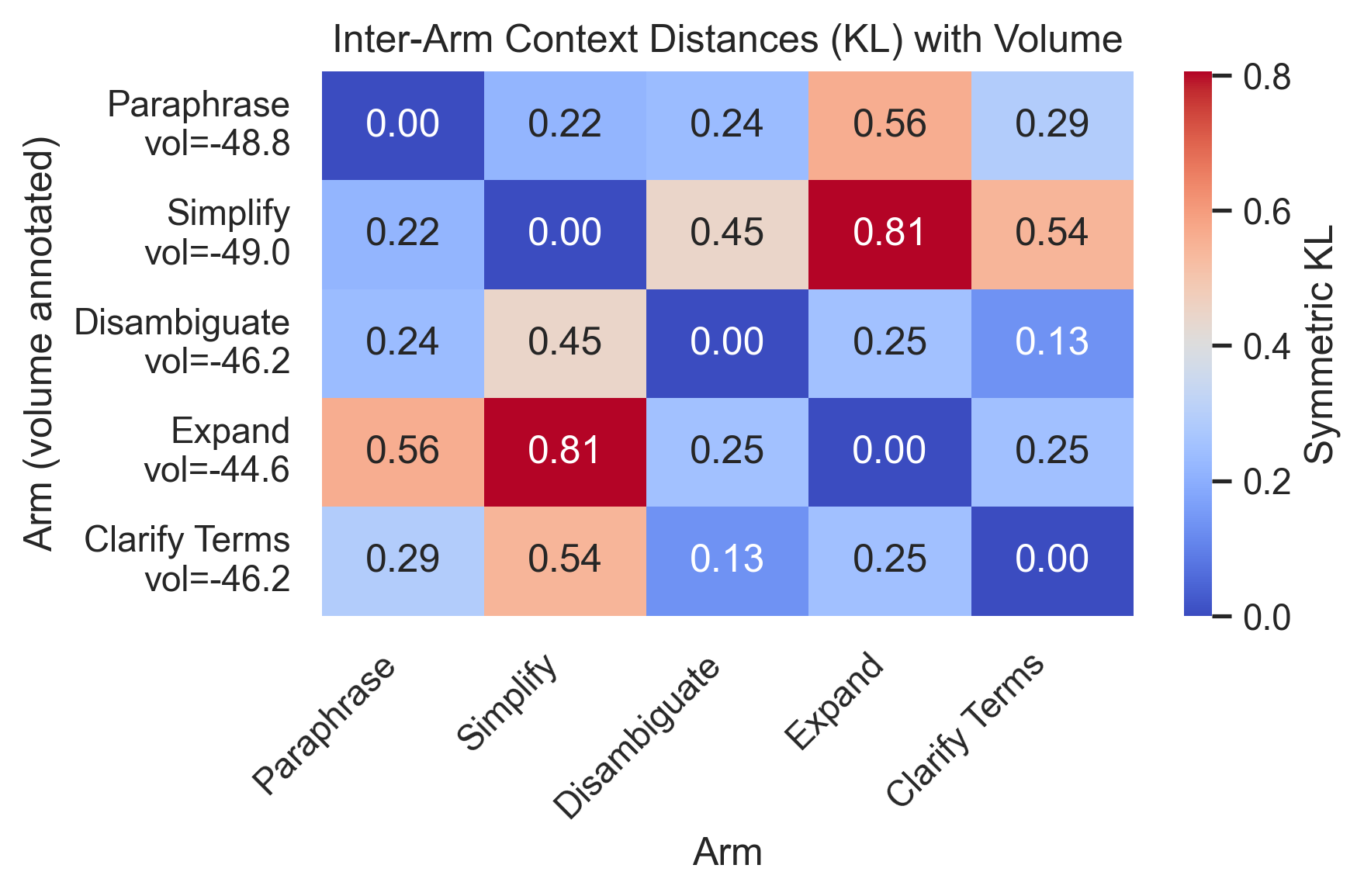}
    \subcaption{Contextual Model KL Distance.}
    \label{fig:feat_kl_lin}
  \end{minipage}\hfill
  \begin{minipage}[t]{0.49\textwidth}
    \centering
    \includegraphics[width=\textwidth]{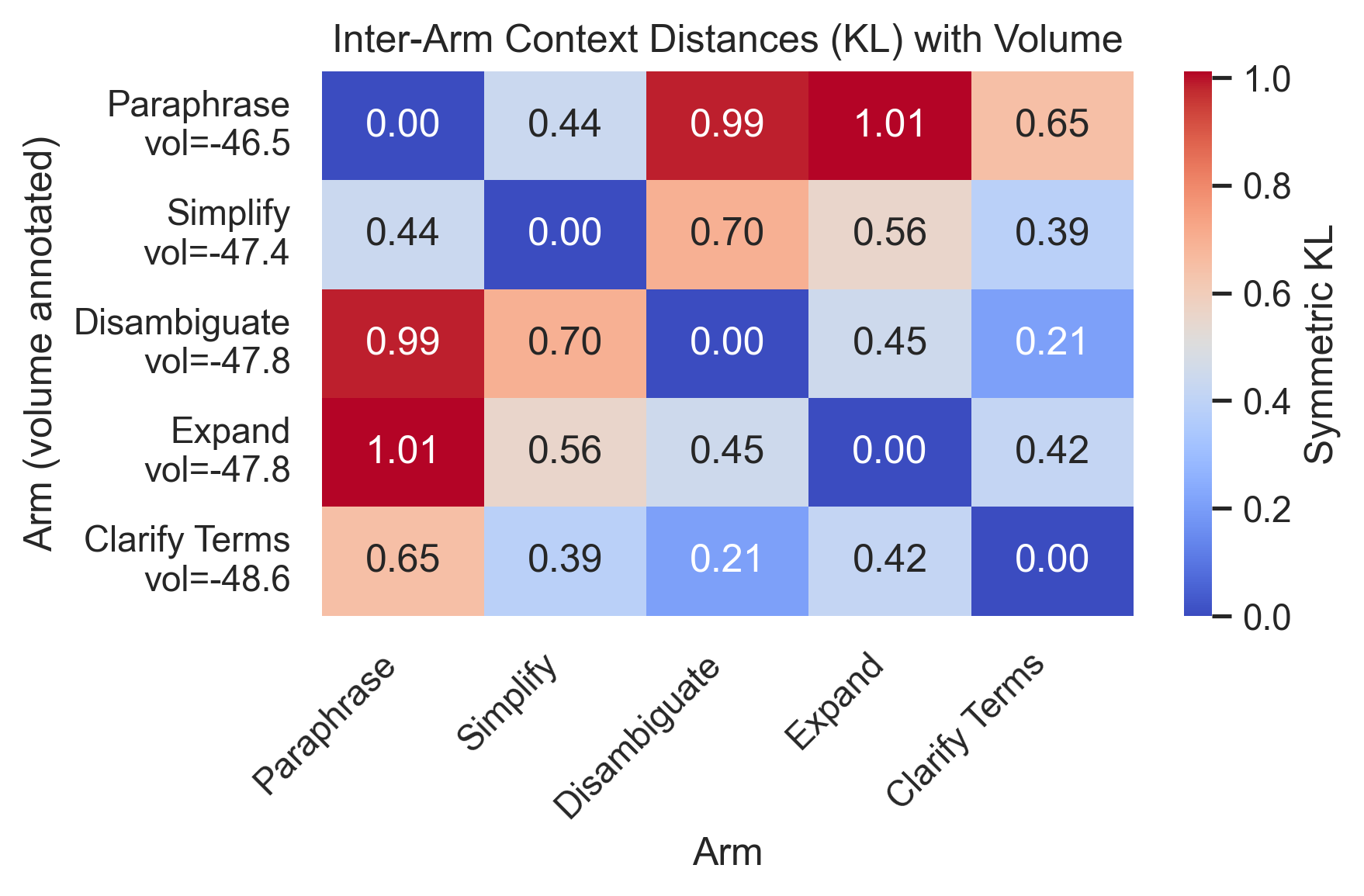}
        \subcaption{Non-Contextual Model KL Distance.}
    \label{fig:feat_kl_nonlin}
  \end{minipage}
  \caption{Comparison of Inter-Arm Context Distances (Symmetric KL) between (\subref{fig:feat_kl_lin}) our contextual bandits and (\subref{fig:feat_kl_nonlin}) its non-contextual counterparts. Arm pairs such as \textsc{Expand} and \textsc{Paraphrase} in the non-contextual bandit setting exhibit high KL distances at 1.01. One interpretation is that the context-clouds barely overlap from dataset to dataset (Figure~\ref{fig:non_lin_diversity}).}
  \label{fig:kl}
\end{figure*}

\begin{figure*}[t]
  \centering
  \begin{minipage}[t]{0.49\textwidth}
    \centering
    \includegraphics[width=\textwidth]{feature_contrib_raw_lin.png}
    \subcaption{Contextual Model Raw Feature Strength.}
    \label{fig:feat_raw_lin}
  \end{minipage}\hfill
  \begin{minipage}[t]{0.49\textwidth}
    \centering
    \includegraphics[width=\textwidth]{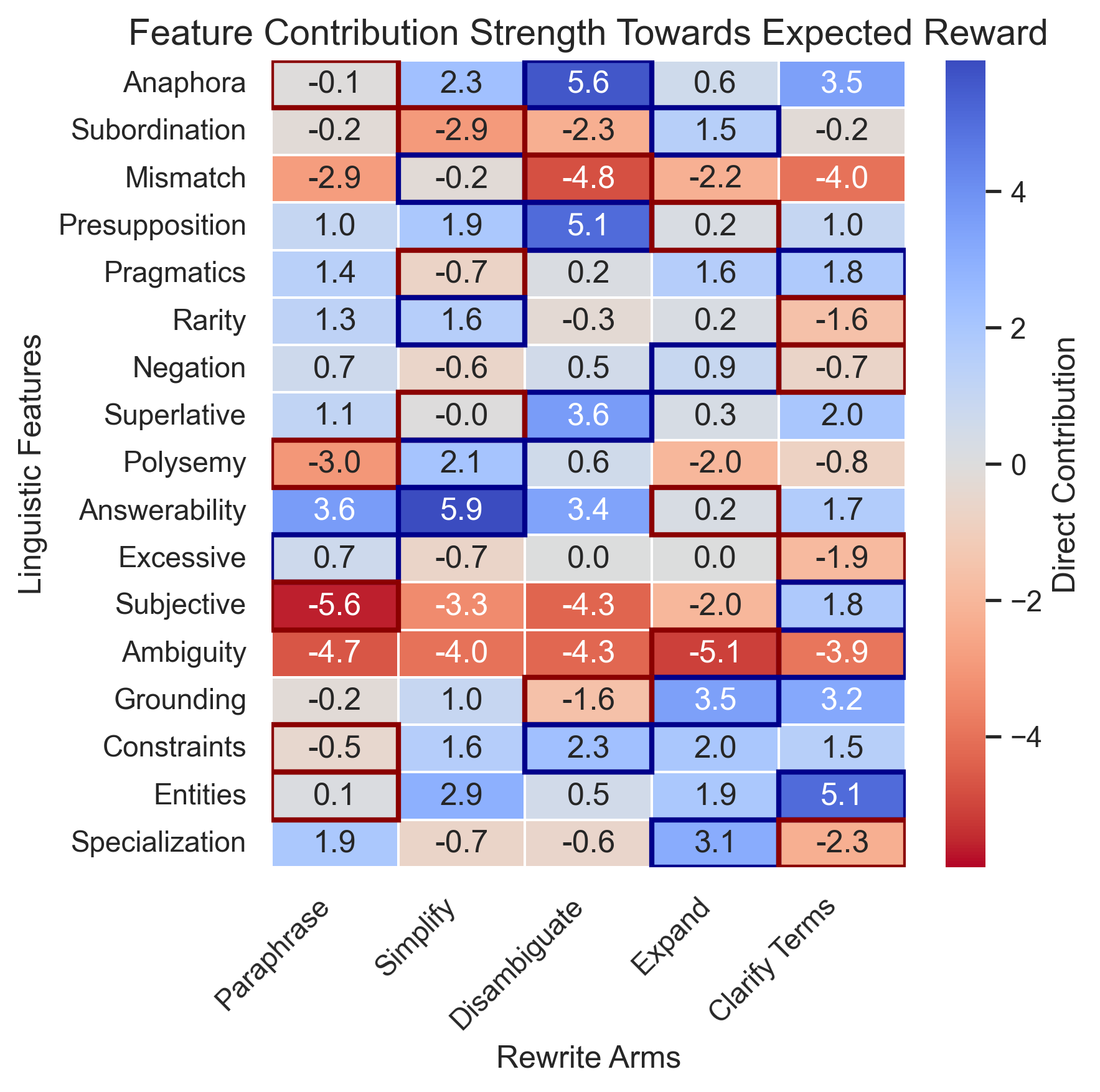}
        \subcaption{Non-Contextual Model Raw Feature Strength.}
    \label{fig:feat_raw_nonlin}
  \end{minipage}
  \caption{Comparison of Raw feature‐level regression coefficients between (\subref{fig:feat_rel_lin}) our contextual bandits and (\subref{fig:feat_rel_nonlin}) its non-contextual counterparts.  Each cell shows how enables a raw view into how specific linguistic feature changes the expected reward under each rewrite strategy.}
  \label{fig:raw}
\end{figure*}

\subsection{FTRL}
The algorithm is initialized with the following parameters: number of arms $n_{\text{arms}}$, dimension $d$, learning rate $\alpha$, exploration parameter $\beta$, and regularization parameters $l_1$ and $l_2$. The cumulative gradient vectors for each arm are stored in $\mathbf{z}_a$, initialized as zero vectors of dimension $d$.

The weight vector $\mathbf{w}_a$ for each arm $a$ is computed as:
$$
w_i = 
\begin{cases} 
-\frac{z_i - \text{sign}(z_i) \cdot l_1}{\frac{\beta + \sqrt{n_i}}{\alpha} + l_2} & \text{if } |z_i| > l_1 \\
0 & \text{otherwise}
\end{cases}
$$

where $z_i$ is the cumulative gradient for the $i$-th feature of arm $a$, and $n_i$ is the cumulative squared gradient for the $i$-th feature. The arm with the highest score, calculated as the dot product of the weight vector $w$ and the context vector, is selected:
$$
a_t = \arg\max_{a \in \{1, \ldots, n_{\text{arms}}\}} \left( \sum_{i=1}^{d} w_i \cdot \mathbf{x}_i \right)
$$

Upon receiving a reward $r_t$ for the selected arm $a_t$, the algorithm updates the cumulative gradient vector $\mathbf{z}$ and the squared gradient sum $\mathbf{n}$ for the selected arm:
\begin{align*}
\varepsilon_{error} &= \langle w, \mathbf{x} \rangle - r_t \\
g &= \varepsilon_{error} \cdot \mathbf{x} \\
\sigma &= \frac{\sqrt{n_i + g_i^2} - \sqrt{n_i}}{\alpha}\\
z_i &\leftarrow z_i + g_i - \sigma \cdot w_i \\
n_i &\leftarrow n_i + g_i^2
\end{align*}

This formulation allows the FTRL algorithm to adaptively adjust the exploration-exploitation trade-off by incorporating both the cumulative reward and the uncertainty in the form of regularization terms, which are scaled by the learning rate $\alpha$ and exploration parameter $\beta$.

\subsection{Linear EXP3}
The algorithm is initialized with parameters: number of arms $n_{\text{arms}}$, dimension $d$, exploration parameter $\gamma$, and learning rate $\eta$. Each arm $a$ maintains a parameter vector $\theta_a$, initialized as $\mathbf{0}_d$.

We compute the probability distribution over arms using the following formulation:
\begin{align*}
\text{logits}_a &= \theta_a^\top \mathbf{x} \\
\text{logits} &= \text{logits} - \max(\text{logits}) \\
\text{exp\_logits}_a &= \exp(\text{logits}_a) \\
\text{base\_probs}_a &= \frac{\text{exp\_logits}_a}{\sum_{a=1}^{n_{\text{arms}}} \text{exp\_logits}_a} \\
\text{probs}_a &= (1 - \gamma) \cdot \text{base\_probs}_a + \frac{\gamma}{n_{\text{arms}}}
\end{align*}

where $\mathbf{x}$ is the context vector. The arm is selected based on the probability distribution $\text{probs}$.

The \texttt{update} method updates the parameter vector $\theta_a$ for the selected arm $a$ using the estimated reward $\hat{r}_t$:
\begin{align*}
\hat{r}_t &= \frac{r_t}{p_a} \\
\theta_a &\leftarrow \theta_a + \eta \cdot \hat{r}_t \cdot \mathbf{x}
\end{align*}
where $p_a$ is the probability of selecting arm $a$, and $r_t$ is the received reward. This strategy leverages exponential weighting and exploration bonuses to balance exploration and exploitation in a linear contextual setting.

\subsection{Linear FTPL}
The algorithm is initialized with parameters: number of arms $n_{\text{arms}}$, dimension $d$, and learning rate $\eta$. Each arm $a$ maintains a parameter vector $\theta_a$, initialized as $\mathbf{0}_d$.

The \texttt{select\_arm} method computes the perturbed scores for each arm using the following formulation:
\begin{align*}
\text{linear\_score}_a &= \theta_a^\top \mathbf{x} \\
\text{noise}_a &\sim \text{Gumbel}(0, \frac{1}{\eta}) \\
\text{score}_a &= \text{linear\_score}_a + \text{noise}_a
\end{align*}
where $\mathbf{x}$ is the context vector. The arm with the highest perturbed score is selected:
$$
a_t = \arg\max_{a \in \{1, \ldots, n_{\text{arms}}\}} \text{score}_a
$$

$$
\theta_a \leftarrow \theta_a + r_t \cdot \mathbf{x}
$$

This strategy leverages random perturbations from a Gumbel distribution to balance exploration and exploitation, allowing the algorithm to explore suboptimal arms while exploiting the accumulated knowledge of their performance in a linear contextual setting.

\subsection{Thompson Sampling}
For a given $\mathbf{x}$, sample $\tilde{\theta}a \sim \mathcal{N}(\mu_a, \Sigma_a)$ and select the arm maximizing:
\begin{equation}
a^* = \argmax_{a \in \mathcal{A}} \mathbf{x}^\top \tilde{\theta}_a.
\end{equation}
Standard Bayesian linear regression updates are then used to update $\mu_a$ and $\Sigma_a$ based on the observed reward $r$.
\begin{equation}
\begin{aligned}
\Sigma_a^{-1} &\leftarrow \Sigma_a^{-1} + \frac{1}{\sigma^2}\,\mathbf{x}\mathbf{x}^\top, \\
\mu_a &\leftarrow \Sigma_a \Bigl( \Sigma_a^{-1}\mu_a + \frac{1}{\sigma^2}\,\mathbf{x}\,r \Bigr).
\end{aligned}
\end{equation}

\begin{figure*}[t]
  \centering
  \begin{minipage}[t]{0.49\textwidth}
    \centering
    \includegraphics[width=\textwidth]{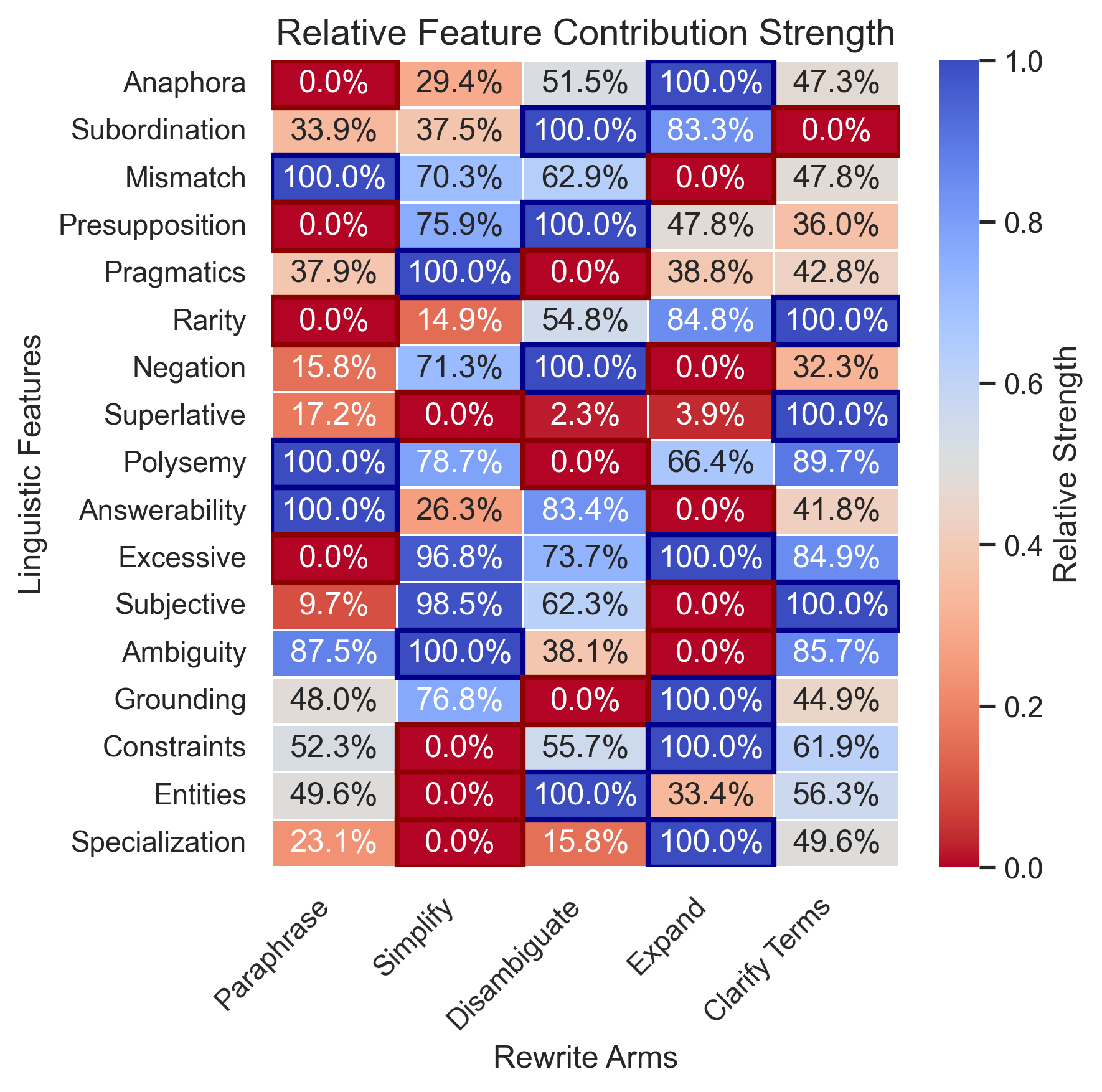}
    \subcaption{Contextual Model Relative Feature Strength.}
    \label{fig:feat_rel_lin}
  \end{minipage}\hfill
  \begin{minipage}[t]{0.49\textwidth}
    \centering
    \includegraphics[width=\textwidth]{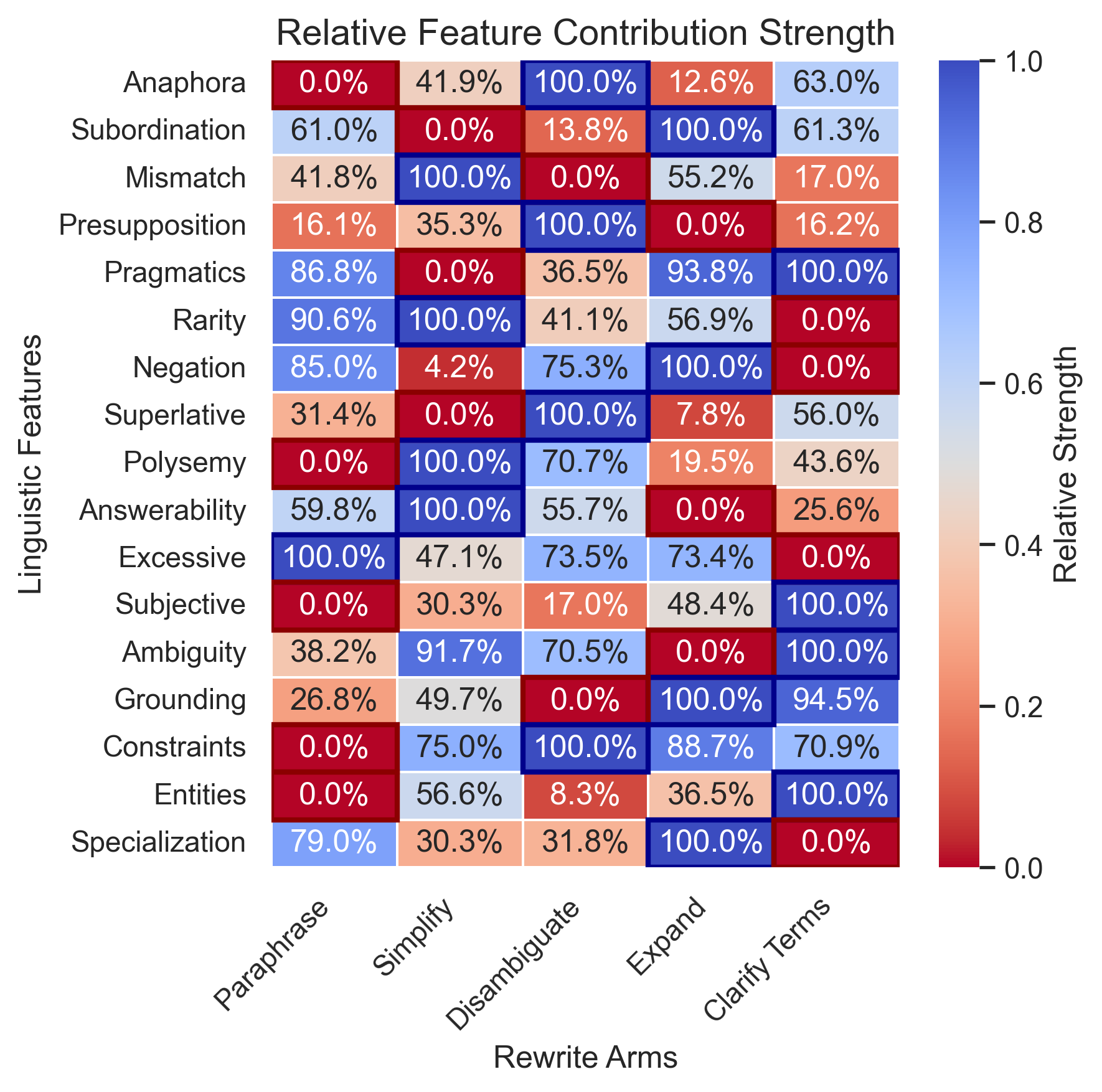}
        \subcaption{Non-Contextual Model Relative Feature Strength.}
    \label{fig:feat_rel_nonlin}
  \end{minipage}
  \caption{Comparison of Min-Max Normalized feature‐level regression coefficients between (\subref{fig:feat_rel_lin}) our contextual bandits and (\subref{fig:feat_rel_nonlin}) its non-contextual counterparts.  Each cell shows how enables a relative view into how specific linguistic feature changes the expected reward under each rewrite strategy. Table~\ref{tab:arm-effects} highlights contextual bandit trends.}
  \label{fig:rel}
\end{figure*}

\begin{table}[t]
  \small
  \centering
    \captionof{table}{%
    \textbf{Top Drivers ($f^+_{\max}$) and Reducers ($f^-_{\max}$) of Rewrite Strategies per Linguistic Features}
    For each rewrite arm, we list the feature whose normalized coefficient
    was highest (100\,\%) and lowest (0\,\%), alongside a brief rationale
    for its positive or negative impact on downstream reward.
    }
  \label{tab:arm-effects}
  \resizebox{\textwidth}{!}{%
    \begin{tabular}{%
        l
        P{2.5cm}
        P{5cm}
        P{2.5cm}
        P{5cm}
      }
      \toprule
      \textbf{Arm $a$}
      & $f^+_{\max}$ 
      & Interpretation
      & $f^-_{\max}$ 
      & Interpretation \\
      \midrule
      \addlinespace[0.5ex]

      \textsc{Disambiguate}
      & Subordination (100\,\%)
      & Long or nested clauses benefit from targeted disambiguation, which isolates and clarifies the core semantic relation.
      & Polysemy (0\,\%)
      & Highly polysemous terms lead disambiguation to pick the wrong sense, degrading downstream reward. \\
      \addlinespace[0.5ex]

      \textsc{Simplify}
      & Pragmatics (100\,\%)
      & Pragmatic cues (e.g.\ discourse markers, politeness) guide safe simplification without loss of meaning.
      & Superlative (0\,\%)
      & Stripping superlative constructions removes essential comparative context, hurting reward. \\
      \addlinespace[0.5ex]

      \textsc{Expand}
      & Constraints (100\,\%)
      & Queries already rich in constraints (time, location, numeric bounds) gain precision when expanded with further qualifiers.
      & Ambiguity (0\,\%)
      & Underspecified queries offer no detail to expand, so further addition of terms only introduces noise. \\
      \addlinespace[0.5ex]

      \textsc{Paraphrase}
      & Answerability (100\,\%)
      & Paraphrasing queries that are already answerable refreshes wording while preserving solvability, boosting LLM performance.
      & Presupposition (0\,\%)
      & Altering queries with strong presuppositions can break implied assumptions, reducing effective reward. \\
      \addlinespace[0.5ex]

      \textsc{Clarify Terms}
      & Rarity (100\,\%)
      & Defining rare or domain-specific terms anchors the LLM’s understanding of technical queries.
      & Subordination (0\,\%)
      & Clarifications in convoluted sentences can introduce further parsing difficulty, impeding reward. \\
      \bottomrule
      \end{tabular}
      }
\end{table}

\begin{figure*}[t]
  \centering
  \begin{minipage}[t]{0.49\textwidth}
    \centering
    \includegraphics[width=\textwidth,
                     trim=0cm 0cm 0cm 0.67cm,
                     clip]{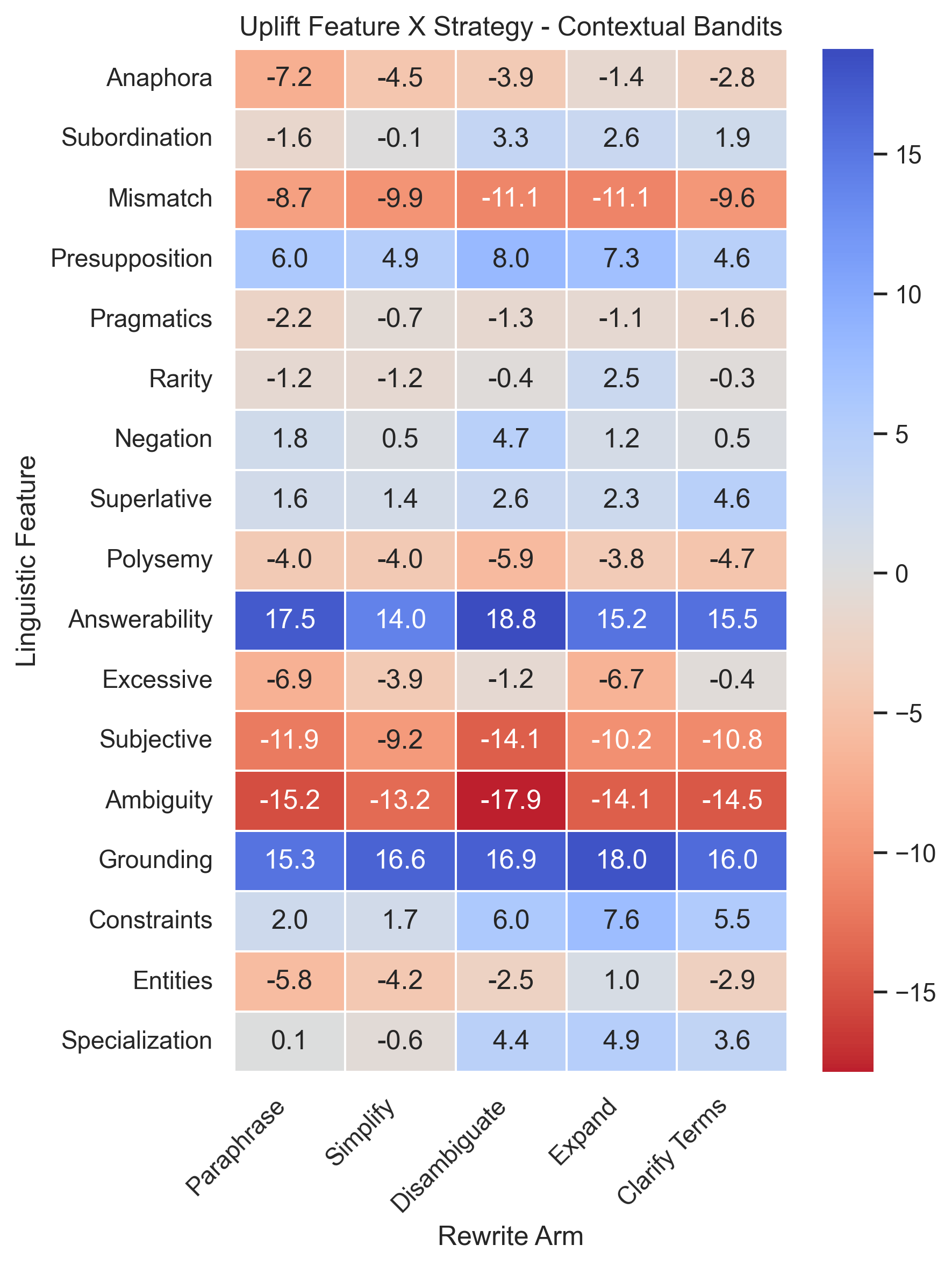}
    \subcaption{Contextual Model Feature Uplift.}
    \label{fig:feature_uplift_linear}
  \end{minipage}\hfill
  \begin{minipage}[t]{0.49\textwidth}
    \centering
    \includegraphics[width=\textwidth,
                     trim=0cm 0cm 0cm 0.67cm,
                     clip]{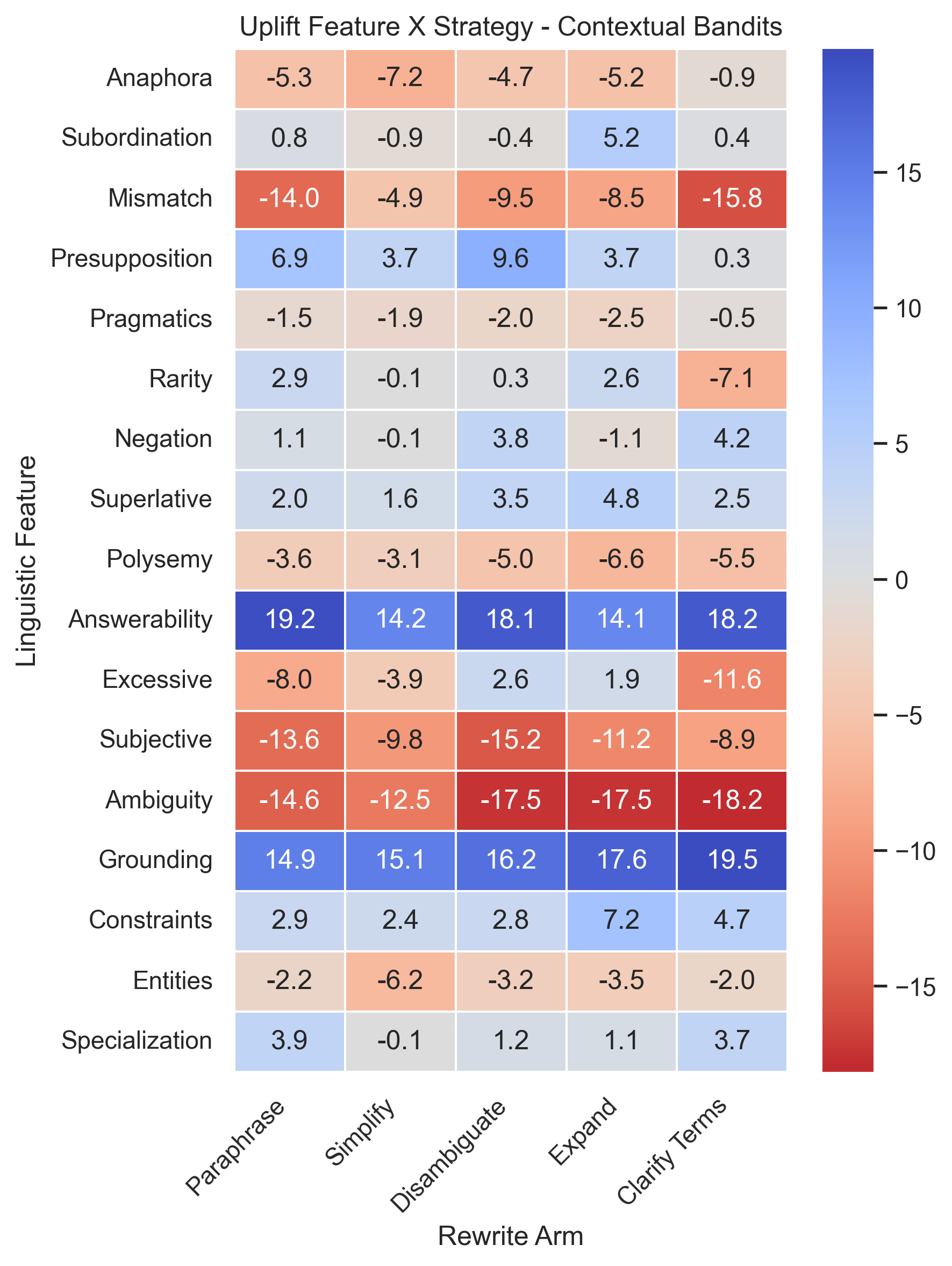}
    \subcaption{Non-Contextual Model Feature Uplift.}
\label{fig:feature_uplift_nonlinear}
  \end{minipage}
  \caption{\textbf{Reward Uplift by Contextual Feature and Strategy.} Feature Uplift measures how much the presence of a binary feature changes the expected reward for a given rewrite arm, formally  
  \(\Delta(f_i,a)=\mathbb{E}[r_t\mid\text{arm}=a,f_i=1]\;-\;\mathbb{E}[r_t\mid\text{arm}=a,f_i=0].\) (\subref{fig:feature_uplift_linear})
  Under the \textbf{contextual} bandit, the strongest positive uplifts come from \textbf{Answerability} ($\approx$ +17 uniformly) and \textbf{Grounding} (+15–18), while \textbf{Ambiguity} ($\approx$ –15 to –18) and \textbf{Subjectivity} ($\approx$ –10 to –14) impose the largest hits across all arms. Mid-range features like \textbf{Presupposition} and \textbf{Constraints} deliver modest boosts ($\approx$ 5), and \textbf{Excessive Details} and \textbf{Anaphora} slightly hurt performance ($\approx$ –5 to –7). (\subref{fig:feature_uplift_nonlinear}) The \textbf{non-contextual} bandit amplifies these trends: \textbf{Answerability} and \textbf{Grounding} remain the top drivers ($\approx$ +18–20), but \textbf{Ambiguity} becomes even more detrimental ($\approx$ –17 to –18), and \textbf{Mismatch} drops to nearly –16 under some arms. Notably, the non-contextual model shows a stronger negative effect for \textbf{Excessive Details} (up to –12) and \textbf{Entities} ($\approx$ –6) than the linear one, suggesting it more sharply penalizes noisy contexts. Together, these heatmaps reveal which linguistic signals each rewrite strategy leverages (or struggles with), and how context vs. context-blind policies weigh them differently.}
  \label{fig:feature_uplift}
\end{figure*}


\begin{figure}[ht]
  \centering
  \label{table: specific feature examples}
  \includegraphics[width=\textwidth]{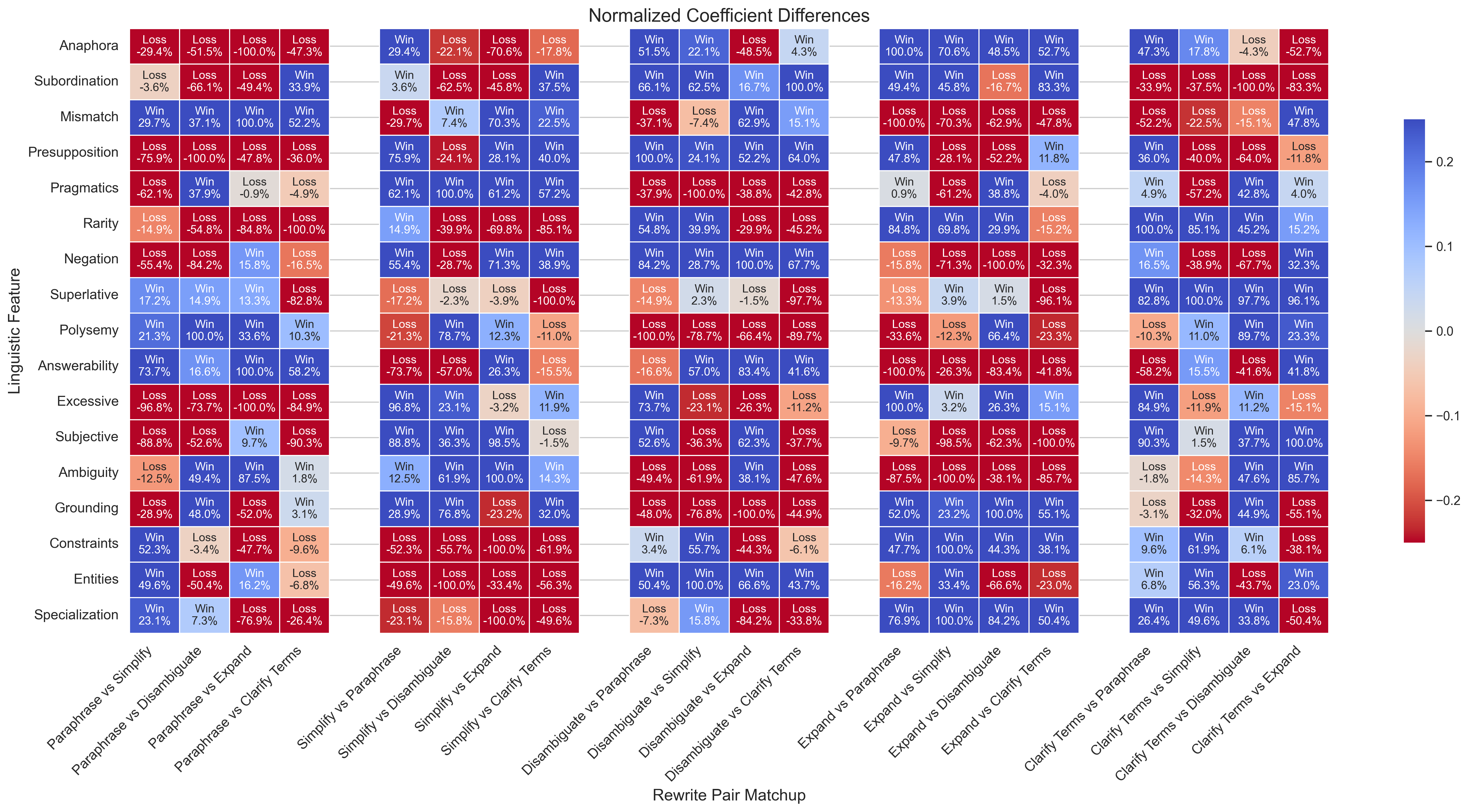}
    \caption{%
    \textbf{Pairwise Normalized Coefficient Differences for Contextual Bandits.}
    Each cell shows the min–max–normalized difference in regression weight
    for a given linguistic feature (rows) between two rewrite arms
    (columns), e.g.\ “Paraphrase vs Disambiguate,” “Simplify vs Expand,” etc.
    Cells labeled “Win” (blue) indicate the feature favors the first arm in
    the matchup, while “Loss” (red) indicates it favors the second. Values
    are expressed as a percentage of the feature’s full coefficient range.}
  \label{fig:coeff-diff-pairwise}
\end{figure}

\begin{table*}[ht]
\small
\centering
\renewcommand{\arraystretch}{1.5}
\setlength{\tabcolsep}{8pt}
\begin{tabular}{clp{5cm}p{4.5cm}}
\toprule
{} & \textbf{Feature} & \textbf{Definition} & \textbf{Example} \\
\midrule

\multirow{3}{*}{\rotatebox{90}{\textbf{Structural}}} 
    & Anaphora & Presence of pronouns or references requiring external context. & "What about that one?" (Unclear reference) \\
    & Subordination & Measures the presence of multiple subordinate clauses & "While I was walking home, I saw a cat that looked just like my friend's." \\
\midrule

\multirow{6}{*}{\rotatebox{90}{\textbf{Scenario-Based}}} 
    & Mismatch & Mismatch between the query’s intended output and its actual structure. & "Find me this paragraph in this document" (When document isn't given, this query cannot be answered) \\
    & Presupposition & Unstated assumptions embedded in the query. & "Who is the musician that developed neural networks?" (Assumes such a musician exists) \\
    & Pragmatics & Captures context-dependent meanings beyond literal interpretation. & "Can you pass the salt?" (A request, not a literal ability) \\
\midrule

\multirow{7}{*}{\rotatebox{90}{\textbf{Lexical}}}
    & Rarity & Use of rare or niche terminology. & "What are the ramifications of quantum decoherence?" (Uses low-frequency terms) \\
    & Negation & Presence of negation words (\textit{not}, \textit{never}). & "Is it not possible to do this?" \\
    & Superlatives & Detection of superlative expressions (\textit{biggest}, \textit{fastest}). & "What is the fastest algorithm?" \\
    & Polysemy & Presence of ambiguous words with multiple related meanings. & "Explain how a bank operates." (Ambiguity: financial institution vs. riverbank) \\
\midrule

\multirow{7}{*}{\rotatebox{90}{\textbf{Stylistic}}} 
    & Answerability & Assesses whether the query has a verifiable answer. & "What is the exact number of galaxies?" (Unanswerable) \\
    & Excessive & Evaluates whether a query is overloaded with information, potentially distracting the model. & "Can you explain how convolutional neural networks work, including all mathematical formulas?" \\
    & Subjectivity & Query requires the degree of opinion or personal bias  & "What is the best programming language?" \\
    & Ambiguity & Highly ambiguous context, task, and wording & "Tell me about history." (Too broad) \\
\midrule

\multirow{8}{*}{\rotatebox{90}{\textbf{Semantic}} }
    & Grounding & Evaluates how clearly the query’s purpose is expressed. & "How does reinforcement learning optimize control in robotics?" (Clear intent) \\
    & Constraints & Identifies explicit constraints (time, location, conditions) provided in the query. & "What was the inflation rate in the US in 2023?" \\
    & Entities & Checks for the inclusion of verifiable named entities. & "Who founded OpenAI?" \\
    & Specialization & Determines whether the query belongs to a specialized domain (e.g., finance, law). & "What are the legal implications of the GDPR ruling?" \\
\bottomrule
\end{tabular}
\caption{Detailed Summary and Examples of Feature Categories, Definitions, and Examples (See Table~\ref{tab:feature_vector_citations})}
\label{tab:feature_examples}
\end{table*}

\end{document}